\pdfoutput=1

\documentclass[11pt]{article}

\usepackage[preprint]{acl}

\usepackage{times}
\usepackage{latexsym}

\usepackage[T1]{fontenc}

\usepackage[utf8]{inputenc}

\usepackage{microtype}

\usepackage{inconsolata}

\usepackage{graphicx}
\usepackage{subcaption}
\usepackage{booktabs}
\usepackage{colortbl}
\usepackage{amsfonts}
\definecolor{LightCyan}{rgb}{0.88,1,1}
\usepackage{tcolorbox} %
\tcbuselibrary{most}
\usepackage[inline]{enumitem}

\title{Assessing the Role of Data Quality in Training Bilingual Language Models}

\author{Skyler Seto, Maartje ter Hoeve\thanks{Equal contribution}, Maureen de Seyssel\footnotemark[1], David Grangier\footnotemark[1] \\
  Apple \\
  \texttt{\{sseto,m\_terhoeve,mdeseyssel,grangier\}@apple.com}}

\begin{document}
\maketitle
\begin{abstract}
    Bilingual and multilingual language models offer a promising path toward scaling NLP systems across diverse languages and users. However, their performance often varies wildly between languages as prior works show that adding more languages can degrade performance for some languages (such as English), while improving others (typically more data constrained languages). In this work, we investigate causes of these inconsistencies by comparing bilingual and monolingual language models. Our analysis reveals that unequal data quality, not just data quantity, is a major driver of performance degradation in bilingual settings. We propose a simple yet effective data filtering strategy to select higher-quality bilingual training data with only high quality English data. Applied to French, German, and Chinese, our approach improves monolingual performance by 2–4\% and reduces bilingual model performance gaps to 1\%. These results highlight the overlooked importance of data quality in multilingual pretraining and offer a practical recipe for balancing performance.
\end{abstract}

\section{Introduction}
\label{sec:intro}
Language models (LMs) exhibit exceptional performance on a number of language understanding and knowledge tasks~\cite{brown2020language,bubeck2023sparks, OpenAI2023GPT4TR}. While much of the effort in training language models is focused solely on English, recent  bi- or multilingual models also incorporate other languages \cite{de2019bertje, martin2019camembert,wei2023skywork, faysse2024croissantllm,lample2019cross,xue2020mt5,le2023bloom,yang2024qwen2}.  In many scenarios, it is beneficial to train a multilingual model as (i) maintaining a separate model for each language can be costly in memory and inference constrained settings, (ii) data relevant to some tasks may only be available in specific languages, and (iii) for many languages, the amount of available high quality data is insufficient for pretraining monolingual language models.

  In contrast to work in the vision domain, which shows scaling data (even noisy or from different domains) improves model performance \cite{sun2017revisiting}, prior work in language modeling shows that multilingual models are prone to degradations relative to monolingual models \cite{chang2024multilinguality,conneau2020unsupervised,xu2024survey}. Fundamentally, training multilingual models requires learning the structure and semantics of each language. Thus, such models may require training for longer \cite{conneau2020unsupervised,chang2024multilinguality} and on better data to reach the same performance as prior work shows that the sample complexity of learning from multi-distribution data grows with the number of distributions \cite{haghtalab2022demand}. Critically, there are two main deficits in prior investigations:

\paragraph{Data quality.} 
While prior work studies how data size impacts performance degradations in multilingual models, they do not study data quality in multilingual models \cite{chang2024multilinguality}. Data quality\footnote{Previous papers define data quality according to a few core principles: (i) fluent language \cite{penedo2023refinedweb,raffel2020exploring}, (ii) long form text \cite{li2023textbooks,yang2024qwen2}, and (iii) informative text with educational content and textbook format \cite{li2024datacomp,penedo2024fineweb}. We discuss more data selection in Section~\ref{sec:rw}.} has already been shown to be an important factor in training high performing English language models \cite{li2024datacomp,li2023textbooks,maini2024rephrasing}. 
Because of this, there is growing interest in model-based filters for curating high quality data in monolingual settings \cite{li2024datacomp,messmer2025enhancing,penedo2024fineweb}. There are many other languages for which training a language model is practical as a reasonably large amount of unfiltered data is available \cite{weber2024redpajama,penedo2fineweb2}.  However, the importance of data quality for training multilingual models and data quality filtering in multilingual settings has received little attention.  
There are several challenges that can arise from filtering high quality data for multilingual pretraining as (i) quality filters may work differently across languages, (ii) the density of high quality data and filtered topics may vary, and (iii) the impact of quality filtering may have a small impact across languages in multilingual settings.

\paragraph{Model and data size} Prior works that study gaps in multilingual model performance typically aim to study performance gaps from training on a large number of languages (typically on the order of hundreds), small data per language, and  smaller encoder-style architectures \cite{conneau2020unsupervised,chang2023multilinguality,xu2024survey}.  In this setting, they refer to the gap as the curse of multilinguality.  However, studying gaps in performance in these settings greatly impacts the evaluations that are feasible, makes it difficult to control training data at scale, and does not control for the fact that training a multilingual model is simply a more challenging task, and may require longer training time or higher capacity to achieve the same performance with the same amount of data.

This work focuses on exploring how these challenges underlie gaps in multilingual performance. We conduct experiments on data quality (measured by information/knowledge) and language in German and Chinese, where we control both through training on translated data.  We also provide a recipe for obtaining high quality data in other languages that improves bilingual model performance in three languages: French, German, and Chinese\footnote{We select these languages for their use in prior work, distance from English, amount of data available, and availability of evaluation benchmarks.}. 
Collectively, our main contributions show that (i) High quality data filtering in multiple languages without access to native high quality data improves performance in the target language, and reduces gaps in monolingual and bilingual performance. (ii) Data quality plays an important role in the performance of bilingual language models (rather than only the language or data size). (iii) High quality English data alone is insufficient for training high performing multilingual language modeling in some tasks.

\section{Related Work}
\label{sec:rw}
\begin{figure*}[ht]
    \centering
       \begin{subfigure}{0.23\textwidth}
        \centering
        \includegraphics[width=\textwidth]{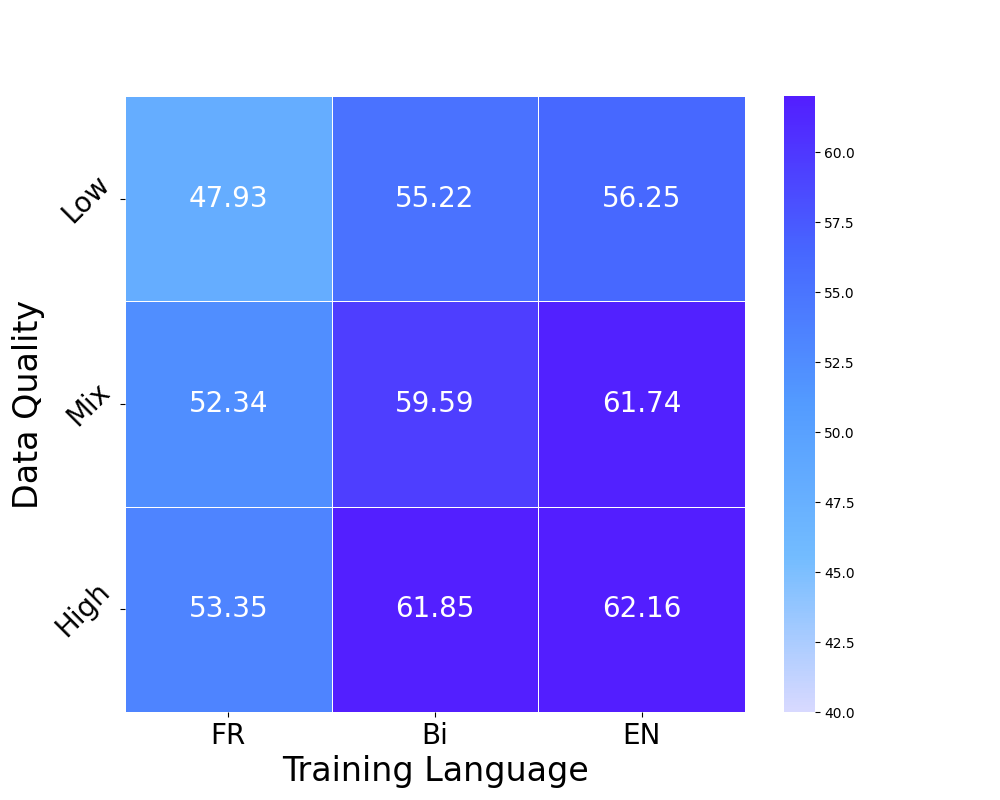}
        \caption{English Core}
        \label{fig:en_fr_core_200}
    \end{subfigure}
    \hfill
       \begin{subfigure}{0.23\textwidth}
        \centering
        \includegraphics[width=\textwidth]{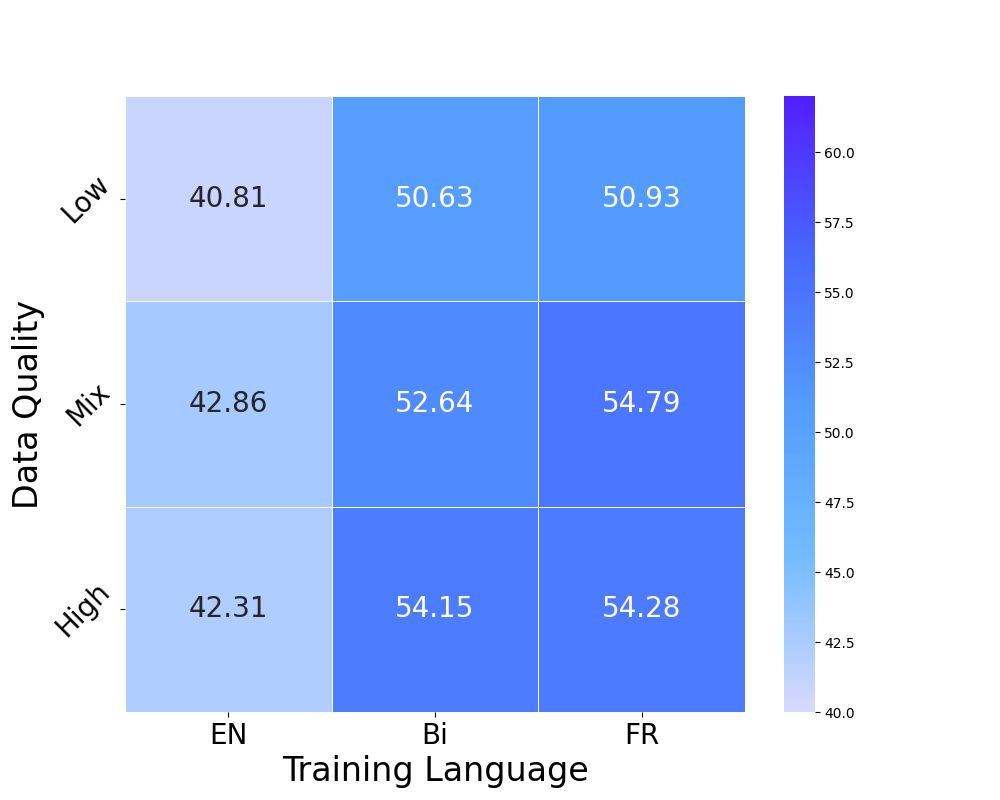}
        \caption{French Core}
        \label{fig:fr_core_200}
    \end{subfigure}
    \hfill
       \begin{subfigure}{0.23\textwidth}
        \centering
        \includegraphics[width=\textwidth]{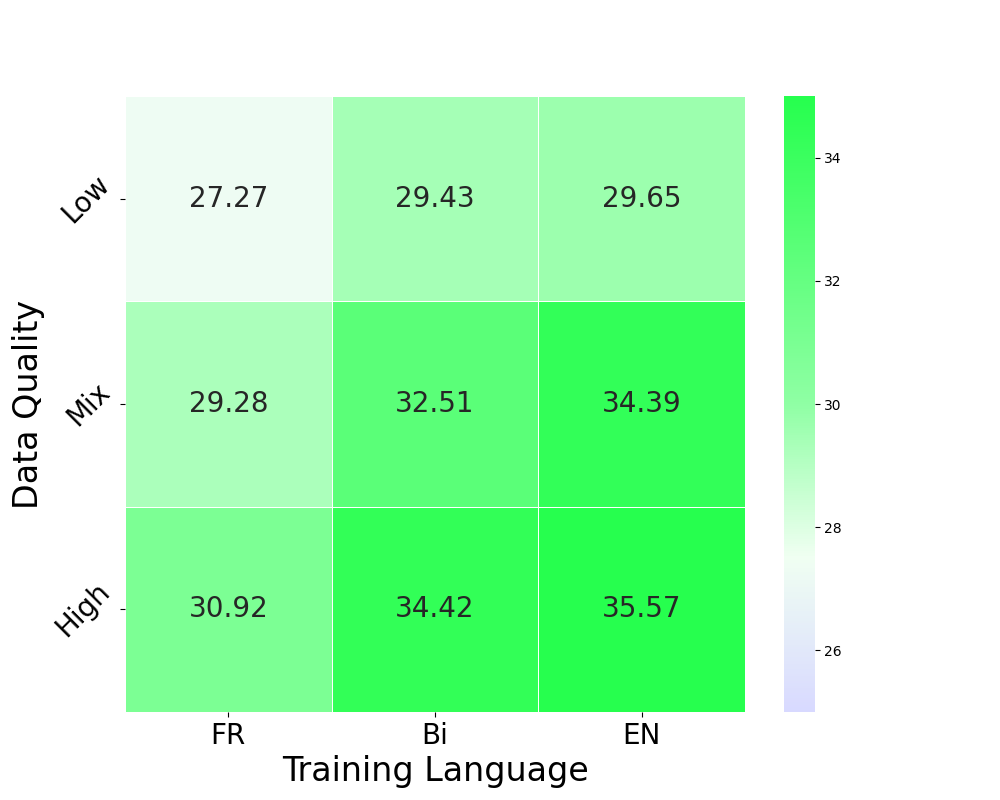}
        \caption{English MMLU}
        \label{fig:en_fr_mmlu_200}
    \end{subfigure}
    \hfill
       \begin{subfigure}{0.23\textwidth}
        \centering
        \includegraphics[width=\textwidth]{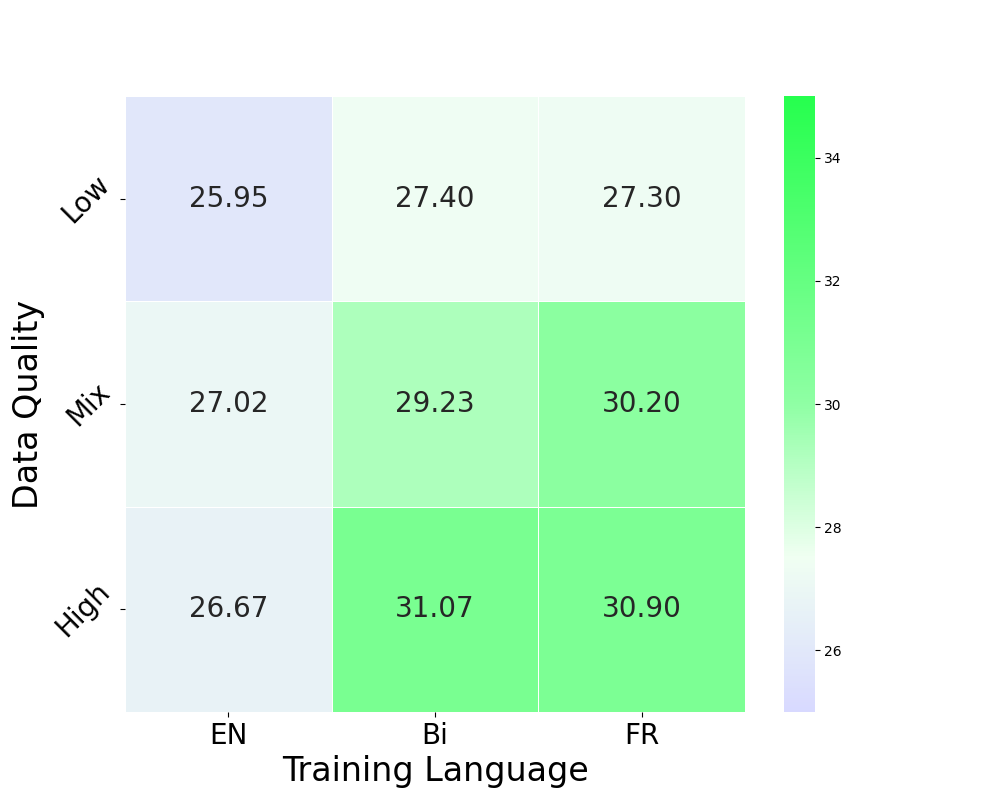}
        \caption{French MMLU}
        \label{fig:fr_mmlu_200}
    \end{subfigure}
       
    \caption{Performance with varying data quality and language.  Models are trained on combinations of mC4 (low) and FineWebEDU (high) in native English (EN) and translated to French (FR). Models are trained for 200K steps and evaluated on Core (avg over six common-sense reasoning tasks) and MMLU. }
    \label{fig:en_fr_heatmaps_200K}
\end{figure*}

\paragraph{Multilingual Language Models}
Large scale multilingual language models are of two main types: (i) They can be trained on a large corpus of multilingual data such as mC4 \cite{xue2020mt5}, CCNet \cite{wenzek2020ccnet}, or FineWeb2 \cite{penedo2fineweb2}  typically covering in the order of 100 languages.  This includes models such as mBert~\cite{devlin2019bert} , XLM~\cite{conneau2020unsupervised}, mT5~\cite{xue2020mt5}, Bloom~\cite{le2023bloom}, etc. (ii) Bilingual language models such as in French \cite{faysse2024croissantllm,le2019flaubert,martin2019camembert}, German \cite{scheible2020gottbert},  Dutch \cite{de2019bertje}, or Chinese \cite{wei2023skywork}, which are typically small, but can be large in the case of Chinese where an abundance of high quality data is present \cite{yu2025opencsgchinesecorpusseries}.  Other models such as the Llama family~\cite{touvron2023llama}, Mistral~\cite{jiang2023mistral}, Gemini~\cite{team2023gemini}, Palm2~\cite{anil2023palm}, and GPT \cite{OpenAI2023GPT4TR} have been shown to have multilingual capabilities, however a majority of their data is English \cite{xu2024survey}. 

\paragraph{Multilingual Data}
Curated datasets are essential to training language models.  Early multilingual datasets include CCNet~\cite{conneau2020unsupervised}, mC4~\cite{xue2020mt5}, and CulturaX~\cite{nguyen2024culturax} all support over 100 languages, though the largest sources of data are English and even other high resource languages such as German, French, Chinese, and Korean contain 10-50$\times$ less data.  Other datasets such as Redpajamav2 contain over 2.5T tokens, but are limited to only a few Indo-European languages \cite{weber2024redpajama}, and still a factor of 7-10$\times$ less than English.  Recently the FineWeb2 dataset was crafted for many languages with the same heuristic filters as the original FineWeb supporting many languages with data for pretraining \cite{penedo2fineweb2}.  Still, there is only a handful of high quality datasets large enough for training language models in select languages like Chinese \cite{yu2025opencsgchinesecorpusseries}, French, German, and Spanish \cite{penedo2fineweb2,messmer2025enhancing}.

\paragraph{Data Selection}
High quality data selection remains an important area of research in training language models. Early research on data selection was based on heuristics including GPT-2~\cite{radford2019language}, Gopher~\cite{rae2021scaling}, C4~\cite{raffel2020exploring}, and RefinedWeb~\cite{penedo2023refinedweb}. Recent works examine model based filters for labeling general high quality data \cite{sachdeva2024train,li2024datacomp}, or textbook quality data \cite{penedo2024fineweb}. While a majority of this work focuses on English data only, a few works have examined filtering in other languages, such as by perplexity \cite{conneau2020unsupervised} or filtering using models trained on high quality specialized data in select languages \cite{messmer2025enhancing}.  Other forms of data selection include reweighting data \cite{grangier2024task,fan2023doge,xie2024doremi} and are shown to have varying degrees of success when applied in bilingual settings with data constraints, but only for English \cite{seto2024training}.

\section{Data Quality and Multilinguality}
\label{sec:data_quality}

We conduct four types of experiments to demonstrate that the performance gap between a bilingual language model and monolingual models is largely due to the data quality and number of training steps - e.g.,  multilingual models require better training and for longer.  We start with demonstrating that there are performance gaps when not controlling for data quality (Section~\ref{sec:dq_not_controlled}).  We then show that training on a translated pretraining corpus in both languages, thereby controlling data quality, 
yields no gap between monolingual and bilingual performance (Section~\ref{sec:dq_controlled}).  Next, we find that at smaller number of training steps, there is a gap between multilingual and monolingual models, and models learn faster with higher quality data (Section~\ref{sec:dq_small_step}). For the experiments controlling data quality, we show results for English and French translated data here, and refer to Section~\ref{sec:heatmap_german} for English and German experiments . Finally, we show that quality also depends on the information available in the data, and that high quality English data with translations alone is insufficient for training high performing bilingual models (Section~\ref{sec:dq_other_data}).   These experiments are done with Chinese and English given the availability of high quality Chinese data for trainng and downstream evaluations similar to MMLU.

\begin{figure*}[ht]
    \centering
       \begin{subfigure}{0.23\textwidth}
        \centering
        \includegraphics[width=\textwidth]{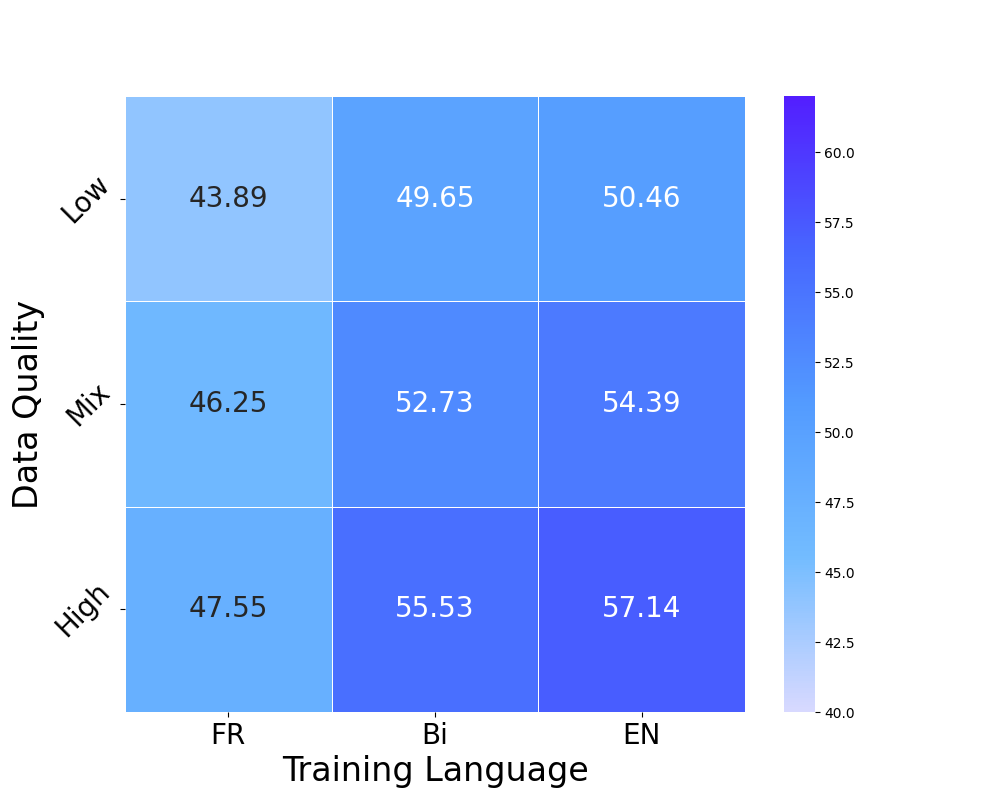}
        \caption{English Core}
        \label{fig:en_fr_core_30}
    \end{subfigure}
    \hfill
       \begin{subfigure}{0.23\textwidth}
        \centering
        \includegraphics[width=\textwidth]{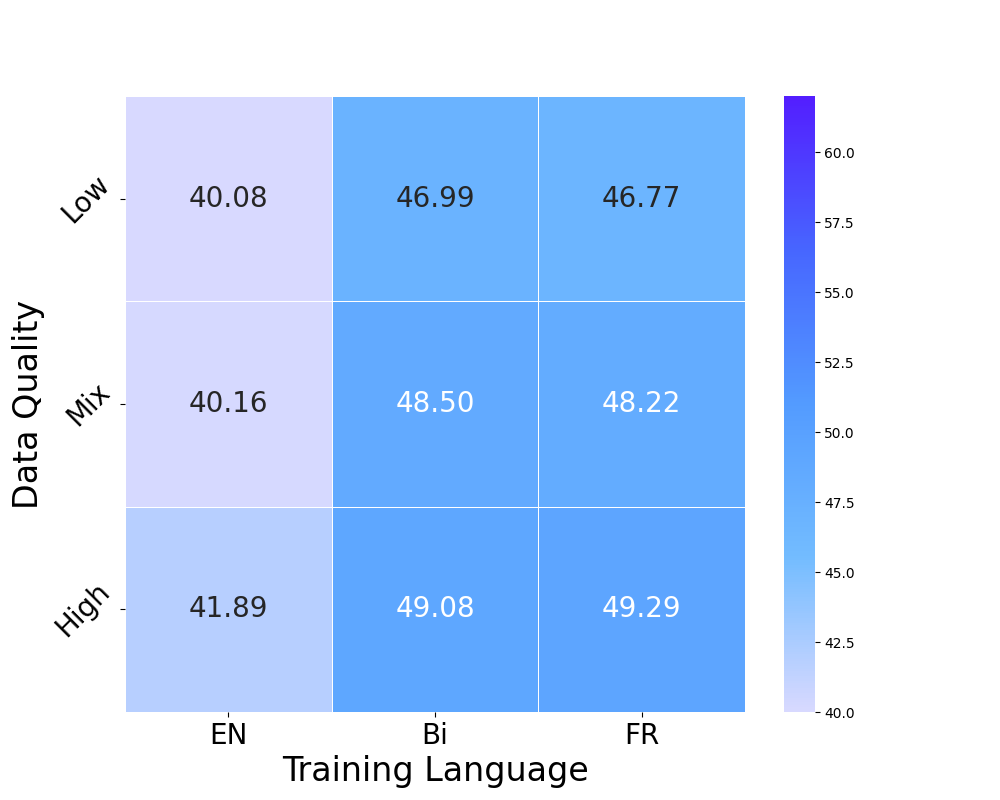}
        \caption{French Core}
        \label{fig:fr_core_30}
    \end{subfigure}
    \hfill
       \begin{subfigure}{0.23\textwidth}
        \centering
        \includegraphics[width=\textwidth]{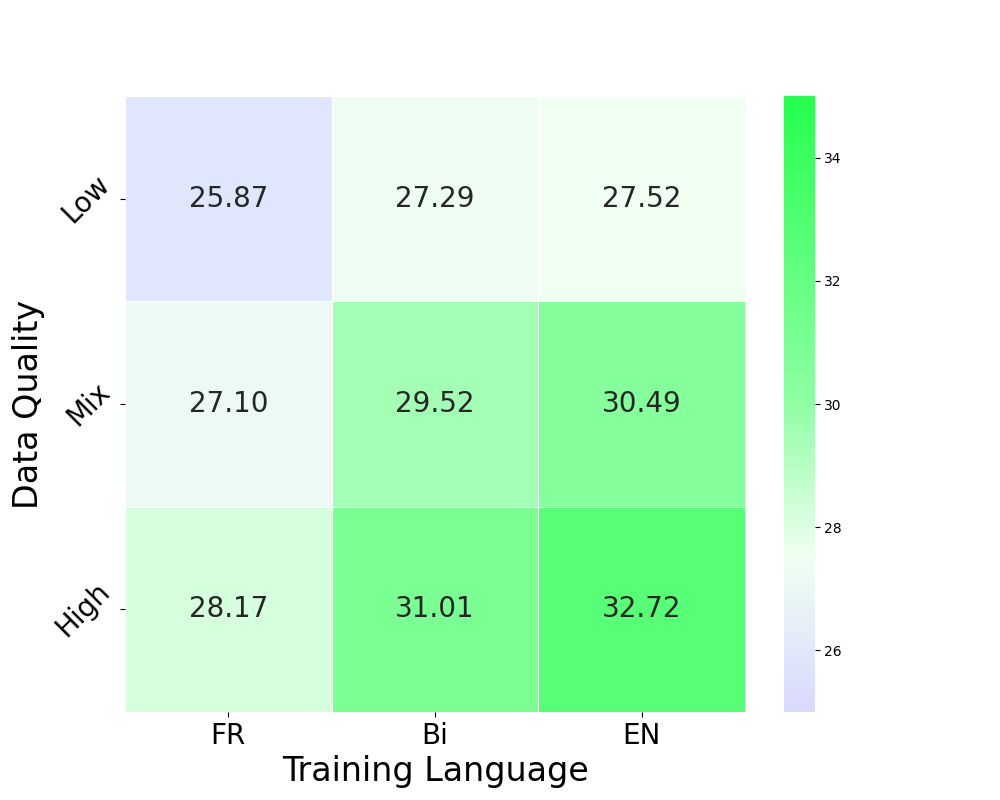}
        \caption{English MMLU}
        \label{fig:en_fr_mmlu_30}
    \end{subfigure}
    \hfill
       \begin{subfigure}{0.23\textwidth}
        \centering
        \includegraphics[width=\textwidth]{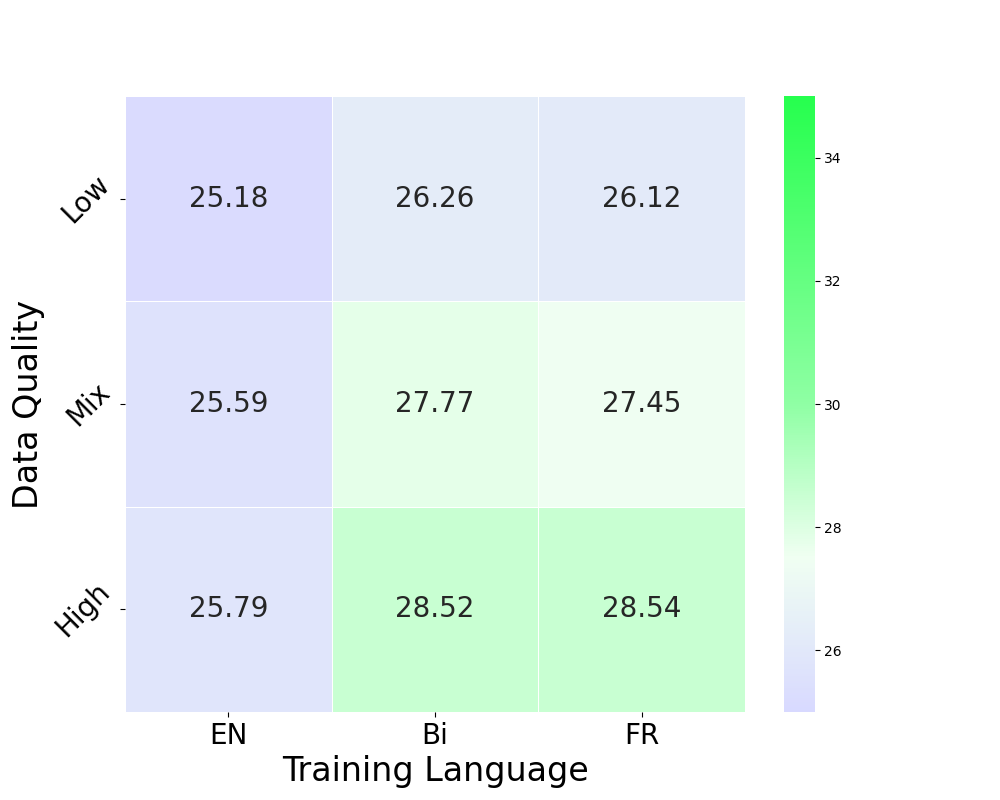}
        \caption{French MMLU}
        \label{fig:fr_mmlu_30}
    \end{subfigure}
       
    \caption{Performance with varying data quality and language.  Models are trained on combinations of mC4 (low) and FineWebEDU (high) in native English (EN) and translated to French (FR). Models are trained for 30K steps and evaluated on Core  and MMLU. }
    \label{fig:en_fr_heatmaps_30K}
\end{figure*}

\subsection{General Model Details}
  We train decoder-only transformer models \cite{vaswani2017attention} with 1.3B non-embedding parameters. Models use the PolyLM tokenizer \cite{wei2023polylm}, with a total vocabulary size of 256K tokens using BPE to allow for using the same tokenizer across all experiments.
 Models are trained for 200K steps  with batch size 1024 and context length 1024 unless otherwise stated. Hyperparameters and model details are in Appendix~\ref{sec:hyperparams}. This model size is chosen as it provides reasonable (above random) performance on several benchmark QA tasks, and is commonly used for benchmarking and ablating pretraining of language models \cite{penedo2023refinedweb,penedo2024fineweb}.

\subsection{Model Performance without Controlled Data Quality}
\label{sec:dq_not_controlled}
\paragraph{Methodology:} We start with a setup in which a bilingual language model is trained on an equal proportion of data from mC4 in  French (FR) and English (EN) totaling 100K steps each.

\begin{table}[ht]
\centering
\resizebox{\columnwidth}{!}{
    \begin{tabular}{lcccc}
        \toprule
        \textbf{Model} & \textbf{Core EN} & \textbf{Core FR} &  \textbf{MMLU EN} & \textbf{MMLU FR} \\
        \midrule
        EN & \textbf{56.3} & 40.8 & \textbf{29.7} & 26.0\\
        FR & 45.9 & 49.0 & 27.0 & 27.2\\
        BI & 53.5 & \textbf{49.6} & 29.3 & \textbf{27.4}\\
        \bottomrule
    \end{tabular}
}
\caption{Zero shot accuracy for general understanding and specialized knowledge tasks for monolingual English (EN), French (FR), and bilingual (BI) models.}
\label{tab:mc4_fr_results}
\end{table}

\paragraph{Findings:} Table~\ref{tab:mc4_fr_results} shows performance on MMLU and Core\footnote{Average over six general knowledge and common-sense reasoning tasks: ARC-easy, ARC-challenge, SciQ, PIQA, HellaSwag, Winogrande} benchmarks.  Our findings match those in prior works including \cite{conneau2020unsupervised,chang2024multilinguality} where we see a 3\% drop in English and an increase in French of 0.8\% compared to the bilingual model for Core tasks. For MMLU, the difference is smaller but the trends remain similar.  The bilingual model has the same ratio of data and sufficient data for training ($\sim 7\times$ Chinchilla scaling).

\subsection{Model Performance Varying Data Quality}
\label{sec:dq_controlled}

\paragraph{Methodology:} To demonstrate that multilingual performance depends on data quality,  we now control the data quality and languages in the model.  We follow the same setup as above and vary both the data quality and the language.  For our experiments, we use the datasets: FineWebEDU, and mC4 EN.  These datasets are chosen as they have 
varying quality according to the DCLM classifier\footnote{A fasttext classifier aimed at distinguishing high quality data according to samples found in OpenHermes and highly upvoted ELI5 posts - \url{https://huggingface.co/mlfoundations/fasttext-oh-eli5}.} (mean quality scores 0.023 vs. 0.1127 respectively), and the FineWebEDU dataset has much higher performance on downstream benchmarks.  

We translate mC4 into French using a proprietary translation system following \cite{seto2024training}, and use TransWebEDU translations for FineWebEDU \cite{wang2025multilingual}. We consider a variety of scenarios where the quality can vary, or the language can vary, and measure the performance on both English and French.

\paragraph{Findings:} Figure~\ref{fig:en_fr_heatmaps_200K} shows the performance difference when varying quality (y-axis) and language (x-axis) for two sets of zero-shot evaluations: Core and MMLU\footnote{We use the continuation version of the task.}. We denote training with the mC4 dataset as low quality, and FineWebEDU as high quality.  When examining the plots, we see that the bottom right square corresponding to a monolingual model\footnote{We flip the x-axis order depending on the evaluation task such that the bottom right is always the high quality monolingual model for consistent comparison.} trained on high quality in the targeted language has the highest performance.  This is closely followed by models which individually vary the language but keep high quality [middle bottom square, e.g., (bi, high)] or mix quality but keep the same language [right middle square, e.g., (EN, mix)], which are all within 1\%. However, mixed quality and mixed language taken together [middle square, e.g., (bi, mix)] exhibits an average 2\% drop in English performance by comparison, compared to each on their own\footnote{Note that the middle square represents the average of two models: as both languages could have the high quality data source}, and all squares with low quality (top row) exhibit a much larger drop than bilingual models with mixed or high quality (bottom four squares).

\subsection{Model Performance with Fewer Steps}
\label{sec:dq_small_step}
\paragraph{Methodology:} Next, we show that training for fewer steps yields a gap between bilingual and native model performance.  Specifically, the experimental setup is the same as above, but we examine training after 30K steps equating to roughly Chinchilla scaling for a 1.3B model.  

\paragraph{Findings:} Results are shown in Figure~\ref{fig:en_fr_heatmaps_30K}.  At this scale, we see that both the bilingual high quality (middle bottom) and mixed quality native monolingual (right middle) models have 2-2.5\% lower performance than the monolingual high quality unlike prior results at 200K steps for English evaluations.  Similarly low quality results (top row) drop below bilingual again indicating data quality has a large role in training.

\subsection{Model Performance with High Quality Data in Multiple Languages}
\label{sec:dq_other_data}
\begin{figure}[ht]
    \centering
       \begin{subfigure}{0.23\textwidth}
        \centering
        \includegraphics[width=\textwidth]{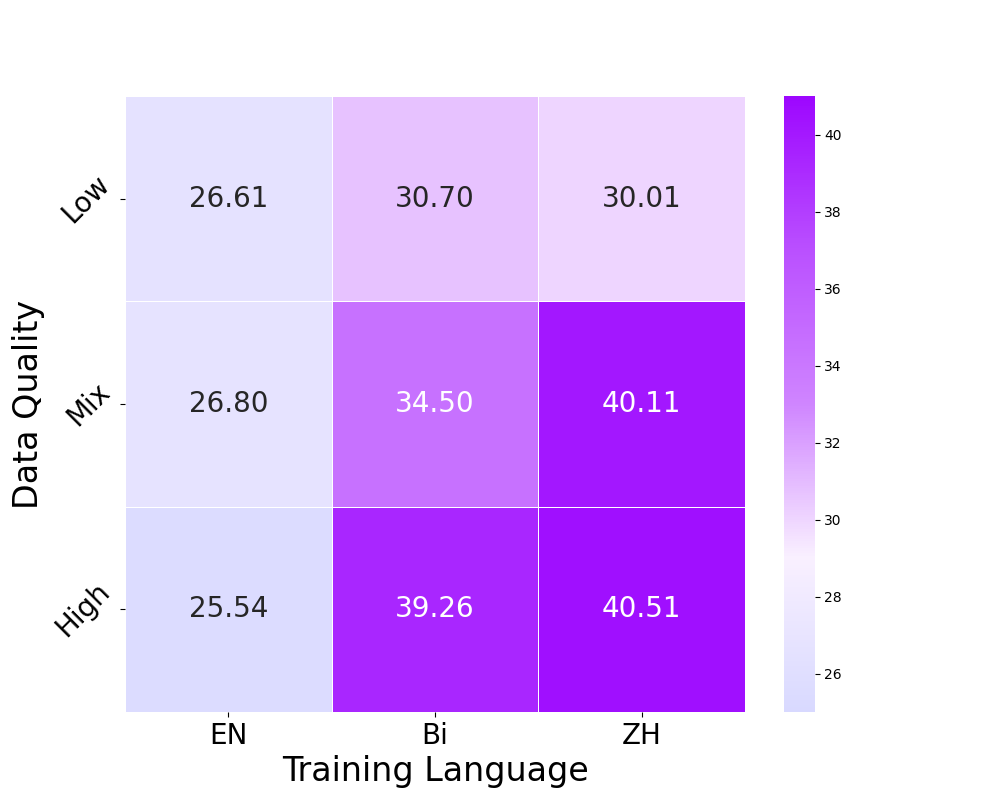}
        \caption{CMMLU ZH}
        \label{fig:zh_cmmlu}
    \end{subfigure}
    \hfill
       \begin{subfigure}{0.23\textwidth}
        \centering
        \includegraphics[width=\textwidth]{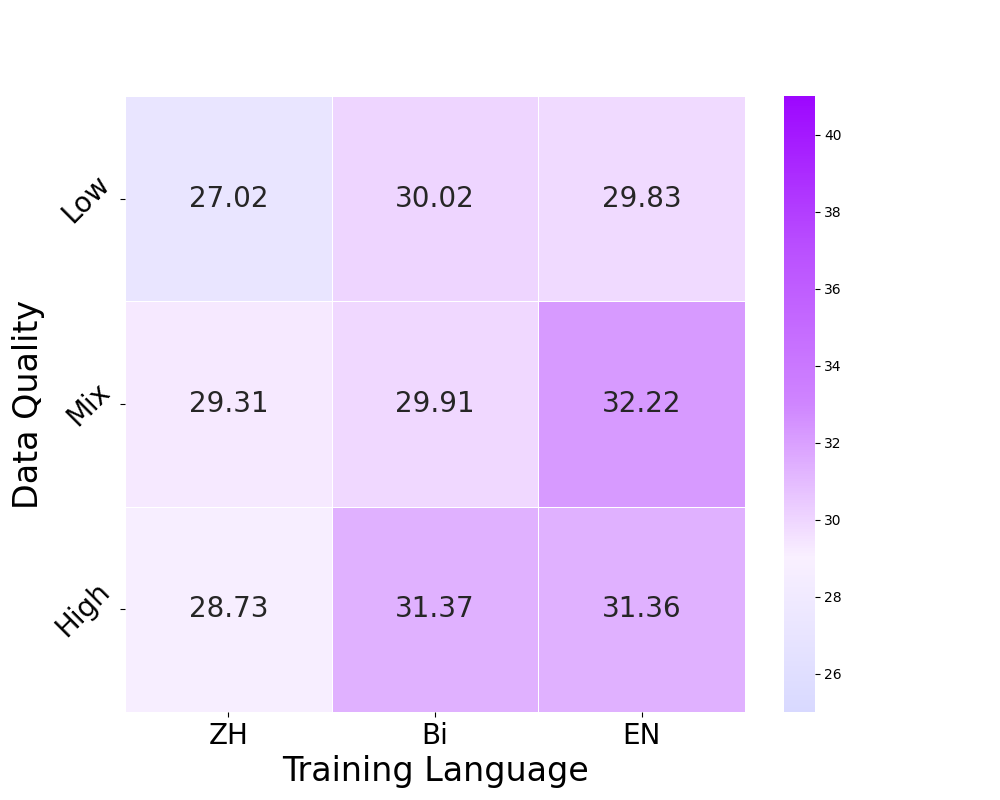}
        \caption{CMMLU EN}
        \label{fig:en_cmmlu}
    \end{subfigure}
       
    \caption{Performance with varying data quality and language.  Models are trained on combinations of Chinese FineWebEDU (high) and FineWebEDU (low) in native English (EN) and translated to Chinese (ZH). Models are evaluated on CMMLU.}
    \label{fig:en_zh_heatmaps}
\end{figure}

\paragraph{Methodology:} We repeat the same training recipe  with two highly curated datasets in different languages: Chinese FineWebEDU and FineWebEDU (which is in English)\footnote{For this experiment, we refer to the FineWebEDU dataset as ``low'' and Chinese FineWebEDU as ``high''.  Here we refer to the drop in quality as (iii) informative text from footnote 1, where FineWebEDU does not cover as much some topics relevant to CMMLU.  This is in contrast to previous sections, for which all definitions of quality drop.}. These datasets are both curated in the same way, and considered high quality in their respective languages, however may contain culturally different information. We translate both datasets into the other language using the same proprietary translation system as in \cite{seto2024training}.  

\paragraph{Findings:} Figure~\ref{fig:en_zh_heatmaps} shows the performance difference for monolingual and bilingual models trained in Chinese and English.  We find that bilingual high quality models (middle bottom square) drop in performance by $\sim 1\%$, but still perform better than low quality Chinese data (right top square).  This drops slightly from the mixed quality monolingual model (right middle square) indicating there may be some effect from further languages.  Nonetheless, our findings show that translated data from English alone may not be sufficient for high quality, and building high quality datasets through the same mechanisms can help yield bilingual models that also perform well in non-translated tasks.

\section{Language-Agnostic Data Filtering}
\label{sec:dq}

\begin{figure*}[ht]
    \centering
       \begin{subfigure}{0.45\textwidth}
        \centering
        \includegraphics[width=\textwidth]{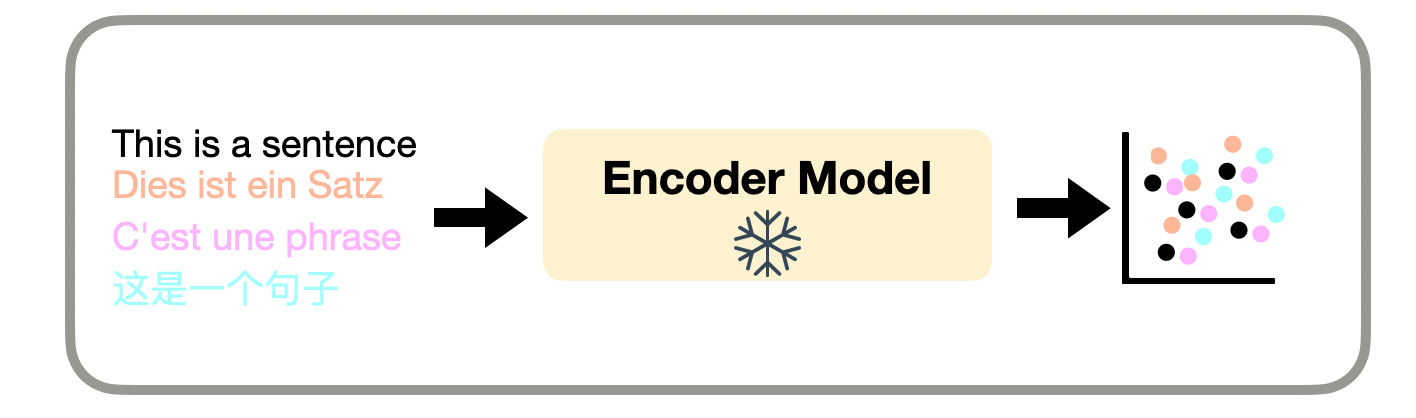}
        \caption{Multilingual Representations}
        \label{fig:sbert_pipeline}
    \end{subfigure}
    \hfill
    \begin{subfigure}{0.45\textwidth}
        \centering
        \includegraphics[width=\textwidth]{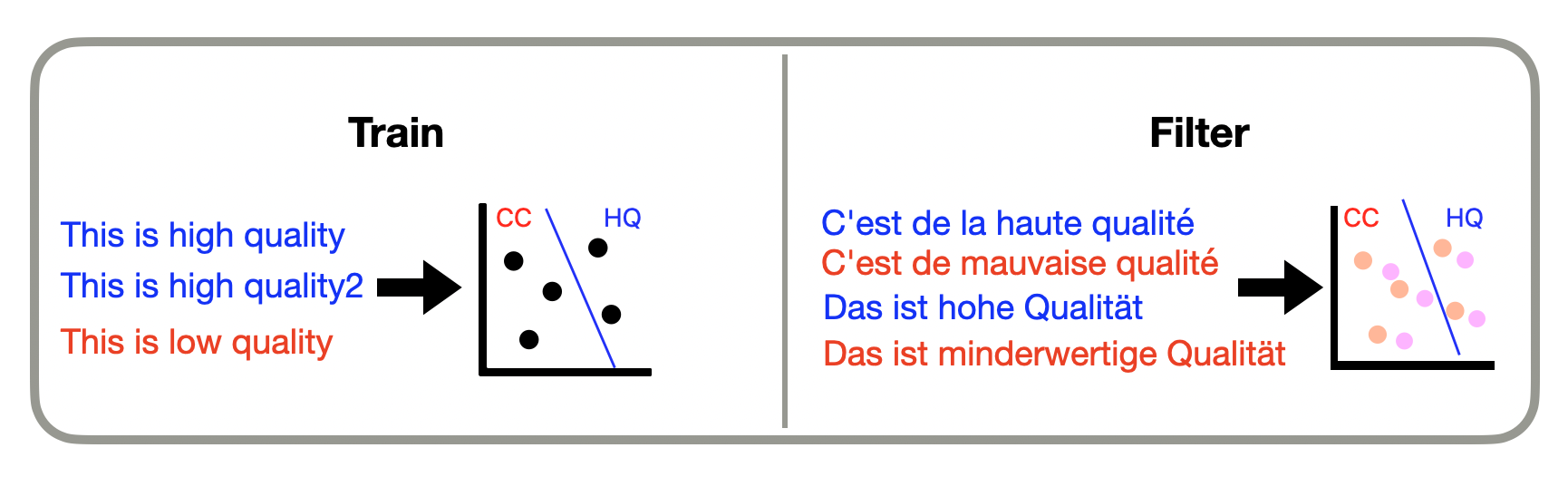}
        \caption{EN Train and Multilingual Filter}
        \label{fig:en_training}
    \end{subfigure}
       
    \caption{(a) \textbf{Multilingual Language Representations}: Build a universal sentence embedding that maps multilingual data to the same embedding space. (b) \textbf{English High Quality Training and Multilingual Data Filtering}: Classifier is trained on the embeddings of a small amount of high quality data available in English. The classifier is then used to filter data in all supported languages.}
    \label{fig:overview}
\end{figure*}

Section~\ref{sec:data_quality} shows that bilingual models may perform as well as monolingual models when the data used to train the models has sufficient information for the target downstream tasks, and is of comparable quality.  However, for a large set of languages, high quality data does not exist.  Prior works have shown that learnable models as quality filters lead to improved performance in downstream tasks for English \cite{li2024datacomp,grangier2024task,penedo2fineweb2}.

To learn a model-based filter for selecting high quality data, we assume access to a small set of high quality data $\mathcal{D}^h = \{(x, y) | x \in \mathbb{R}^d, y \in \{0, 1\}\}$, where $x$ is some representation of a document, and $y$ is a binary label indicating the quality of the sample. A binary classifier $\phi$ is trained on $\mathcal{D}^h$ to estimate the probability that a document from the general pretraining data $\mathcal{D}^g$ is high quality.  The high quality pretraining set $D^{hq}$ is then selected according to the classification rule
\begin{equation}
    \mathcal{D}^{hq} = \{x \in \mathcal{D}^g | \phi(x) > \tau\},
\end{equation}
for some predefined threshold $\tau$.  Unlike prior works which assume training the classifier and selecting general high quality data \cite{li2024datacomp}, or specialized data relevant to downstream tasks \cite{grangier2024task} in English,  we assume the high quality data sample is available only in English, but will be used to select data in other languages.  Concurrent to our work, \cite{messmer2025enhancing} show that a classifier approach in \cite{grangier2024task} can be applied to languages other than English.  They follow a similar setup using specialized datasets, such as translated MMLU \cite{singh2024global}, and Include \cite{romanou2024include}, for data selection.  As such it is still unclear whether a universal language embedding with high quality seed data available only in English can be used to train a language agnostic filter, and whether monolingual filtering improves bilingual model training.  We provide preliminary experiments indicating similar distributions of English and translated French data in Appendix~\ref{sec:sbert_classifier_details}.

In our experiments, we parameterize $\phi$ as a logistic regression, and use a lightweight SentenceBERT (SBERT) multilingual model\footnote{The model, {\tt paraphrase-multilingual-MiniLM-L12-v2}, is at \url{https://huggingface.co/sentence-transformers/paraphrase-multilingual-MiniLM-L12-v2}.} for extracting features \cite{reimers-2019-sentence-bert}.  For $\mathcal{D}^h$, we use the same English data used for training the DCLM classifier, which compares data from RefinedWeb (low quality), and OpenHermes 2.5 or ELI5 (high quality). In our ablations, we also explore using the annotations for FineWebEDU with binary classification as whether the score was above $2.$ The value of $\tau$ is selected to ensure enough data for pretraining, and is in the order of $10\%$ of the data following \cite{li2024datacomp}. In this work, we train a classifier as we have limited data within each cluster for training a 1.3B model at the desired scale, and would repeat data significantly if we instead do data selection for specialized data.

\section{Experiments}
\label{sec:experiments}
\subsection{Experimental Setup Details}
\label{sec:general}
 We use the same 1.3B parameter models trained for 200K steps as in Section~\ref{sec:data_quality} and train with FineWeb2 and Redpajama2 datasets for filtering.  Additional details on exact pools of data are available in Appendix~\ref{sec:appendix_datasets} and different model sizes in Section~\ref{sec:model_scaling}.  We give our main experiments and ablations on English-French bilingual pretraining and include German and Chinese filtering in Section~\ref{sec:german_chinese_filter_eval}. Additional motivation for language selection is in Appendix~\ref{sec:appendix_datasets}. Individual task accuracy in Appendix~\ref{sec:indiv_task_accs}.

\begin{table*}
    \centering
\begin{tabular}{l|cc|ccccc>{\columncolor{LightCyan}}c}
    \toprule 
    \textbf{Model} & \textbf{Core EN} & \textbf{MMLU EN} &  \textbf{Core} & \textbf{MMLU} & \textbf{FB-MC} & \textbf{Regional} &  \textbf{NLI}  & \textbf{AVG}\\
    \midrule
    RPJ2 1.3B & 59.74 & 33.25 &  52.16 & 29.35 & 60.24 & 31.99 & 39.72 &  42.69\\ %
    FW2 1.3B & 60.26 & 33.44 &  53.65 & 29.57 & 60.24 & 31.5 & 39.78 & 42.95 \\
    TWE 1.3B  & 61.85 & 34.43 &  54.15 & \underline{31.07} & 54.48 & 29.55 & 42.14 &  42.28\\ %
    \midrule 
    TranswebLLM & 59.61 & 34.26 & 55.05 & 30.44 & 53.00 & 29.64 & 40.39 & 41.70\\
    CroissantLM & 56.67 & 30.59  & 52.58  & 28.63  & 58.89  & 30.17 & 41.33  & 42.32 \\
    Bloom 1.1B & 51.73 & 29.10 & 48.70 & 27.23 & 55.16 & 28.91 & 40.11 & 40.02\\
    Qwen2.5 1.5B & 69.45 & 41.14 & \textbf{55.65} &  \textbf{32.49} & 59.14 & 31.55 & \underline{43.55} & \textbf{44.50}\\
    EuroLLM 1.7B & 63.05 & 35.39 & 54.94 & 30.20  & \underline{60.34} & 32.54 & 40.42 & 43.69\\
    \midrule
    RPJ2 90\% 1.3B & 59.73 & 33.56 & 53.37 & 29.70 & 59.81 & \textbf{32.91} & \textbf{43.59} & 43.88\\ %
    FW2 90\% 1.3B &  61.07 & 33.54 & \underline{55.47} & 30.16 & \textbf{61.67} & \underline{32.82} & 40.52 & \underline{44.13}\\ 
    \bottomrule
\end{tabular}
\caption{Comparison of different bilingual models in French and English compared with other public multilingual models of similar sizes on French evaluation tasks.}
\label{tab:fr_comp_public}
\end{table*}

\subsection{Monolingual Data Filtering Performance}
\label{sec:dq_perf}
\paragraph{Methodology:} We  show that increasing quality according to our model-based filter leads to an increase in downstream task performance in monolingual settings.  For our experiments, we train a 1.3B parameter model for 30K steps, and evaluate the zero-shot accuracy on the Core set of tasks.  We compare the SBERT filtering classifier with training a model on raw data from RedPajama2, and FineWebEDU in the respective language, as well as filtering RedPajama2, and using a fasttext classifier for filtering trained on the DCLM classifier training data data translated in French. We use two translation systems to show the effect of translation: a cheap proprietary CPU translation system, and the Mistral-7B model following \cite{wang2025multilingual}. We show results for full comparisons for 30K steps (Figure~\ref{fig:sbert_filter_fr}), and for 200K steps in Appendix~\ref{sec:filter_200K}.  %

\begin{figure}[ht]
    \centering
    \includegraphics[width=0.75\linewidth]{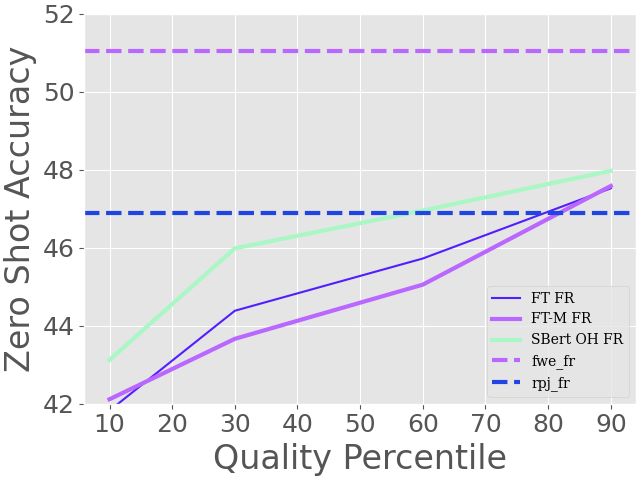}
    \caption{Quality vs. accuracy on Core tasks for filtered RedPajama2 (SBert OH FR) compared with filtering using a FastText classifier (FT FR and FT-M FR), TransWebEDU (fwe\_fr) and base RedPajama2 (rpj\_fr) in French after 30K steps.}
    \label{fig:sbert_filter_fr}
\end{figure}

\paragraph{Findings:} As we increase the percentile of data quality, the accuracy increases leading to a 6\% increase between the lowest and highest quality.  Second, filtering with SBERT outperforms training without filtering for  RedPajama2, and outperforms a fasttext classifier trained on translated data as in \cite{li2024datacomp}, even with a high quality translation system such as Mistral-7B.  Training on the high quality filtered data falls short of the TransWebEDU data, however this is expected to be an upper bound since the evaluations are also translated benchmarks, for which the FineWebEDU data is more highly curated than RedPajama2.

\subsection{Filtering from Already Curated Data}
\label{sec:dq_fw2}
\paragraph{Methodology:} Next, we show that our method also selects high quality data in more highly curated datasets.  We run the filtering on the FineWeb2 French dataset, which has additional heuristic filtering as originally done for the FineWeb2 English data \cite{penedo2fineweb2}, and compare with RedPajama2 French data. We select the top 10\% of data for training for both datasets, noting that the amount of data available in FineWeb2 could be around 10\% of the amount of data in RedPajama2 as estimated by the number of words in the corpus..

\begin{table}[ht]
\centering
\begin{tabular}{lccc}
    Model & RPJ2 & FW2 & TWE \\
    \hline
    Base & 48.38 & 51.61 & 54.28\\
    90\% Filter & 50.42 & 54.17 & -- \\
    \hline
\end{tabular}

\caption{Performance on Core French benchmarks comparing monolingual models with and without filtering across RedPajama2 (RPJ2), FineWeb2 (FW2), and TransWebEDU French (TWE) datasets.}
\label{tab:fw2_rpj_fr}
\end{table}

\paragraph{Findings:} Table~\ref{tab:fw2_rpj_fr} shows results for monolingual French models.  Our results indicate that performance increases even with the smaller FineWeb2 (FW2) dataset and repeated epochs of training. We find that many of the heuristic filters and text extraction also lead to better performance as the base FineWeb2 improves on even the 90\% filter over the Redpajama2 (RPJ2) data.  Finally, we note that the performance of the filtered FineWeb2 data matches that of TransWebEDU indicating similar performance to highly curated translated English data on translated evaluations. Experiments for other percentiles on FineWeb2 are in Appendix~\ref{sec:additional_filter}.

\begin{figure}[ht]
    \centering
    \includegraphics[width=\linewidth]{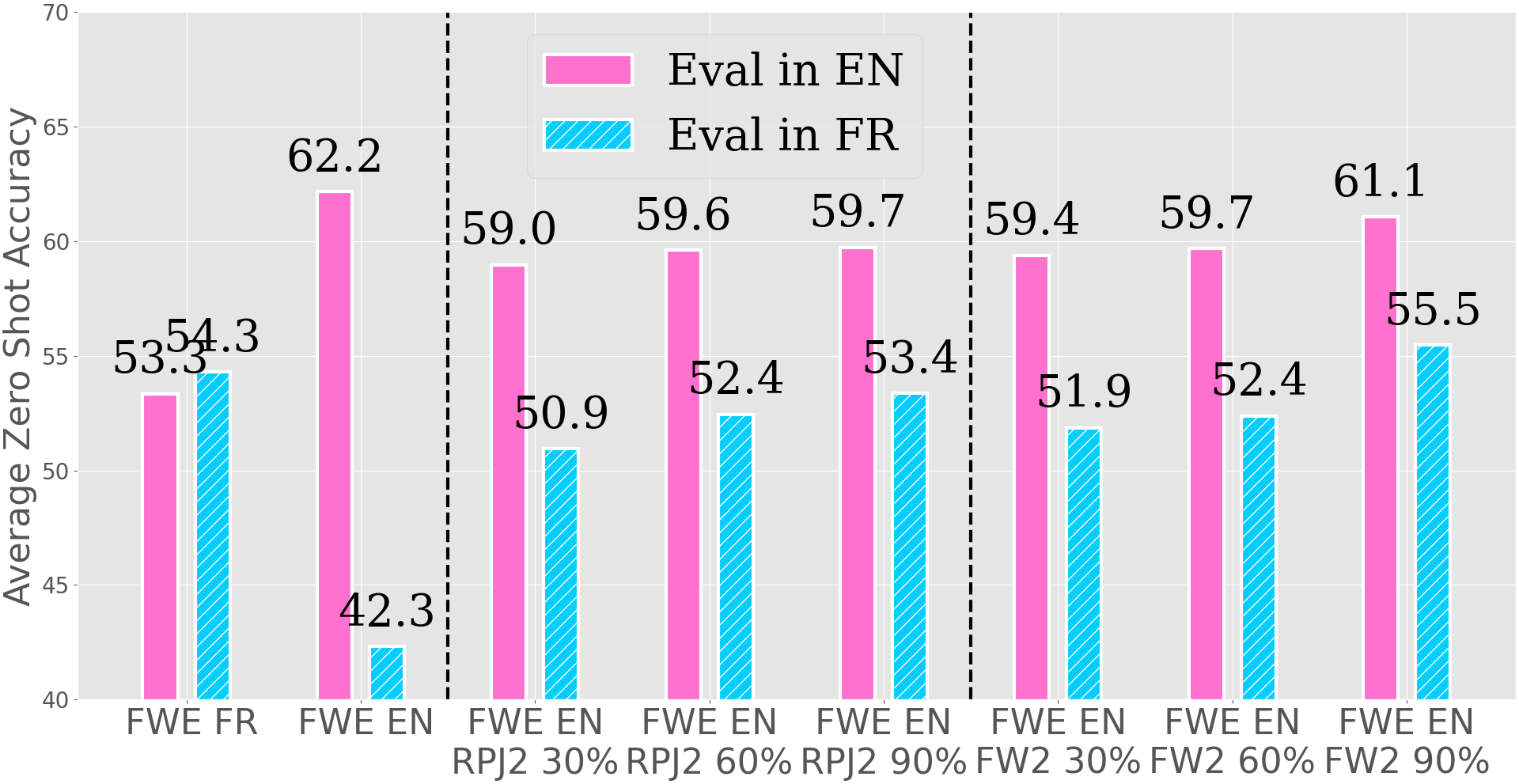}
    \caption{Bilingual vs. Monolingual performance on Core English (EN) and French (FR) benchmarks with Filtering in French. All models use the TransWebEDU English (FWE EN) data while varying the French data. }
    \label{fig:filter_bilingual}
\end{figure}

\subsection{Filtering for Bilingual Models}
\label{sec:bilingual_data_quality}

\paragraph{Methodology} Now, we show in addition to monolingual performance gains, our data selection method diminishes the gap in bilingual models matching the lack of performance gap we see in Section~\ref{sec:data_quality} with native data.  For our experiments we compare models trained on data at different quality percentiles at the 30th, 60th, and 90th percentile, and take a total of roughly 10\% of the data. We evaluate on the Core evaluations and report results in Figure~\ref{fig:filter_bilingual} comparing monolingual TransWebEDU performance with bilingual models.

 \begin{figure*}[ht]
    \centering
       \begin{subfigure}{0.3\textwidth}
        \centering
        \includegraphics[width=\textwidth]{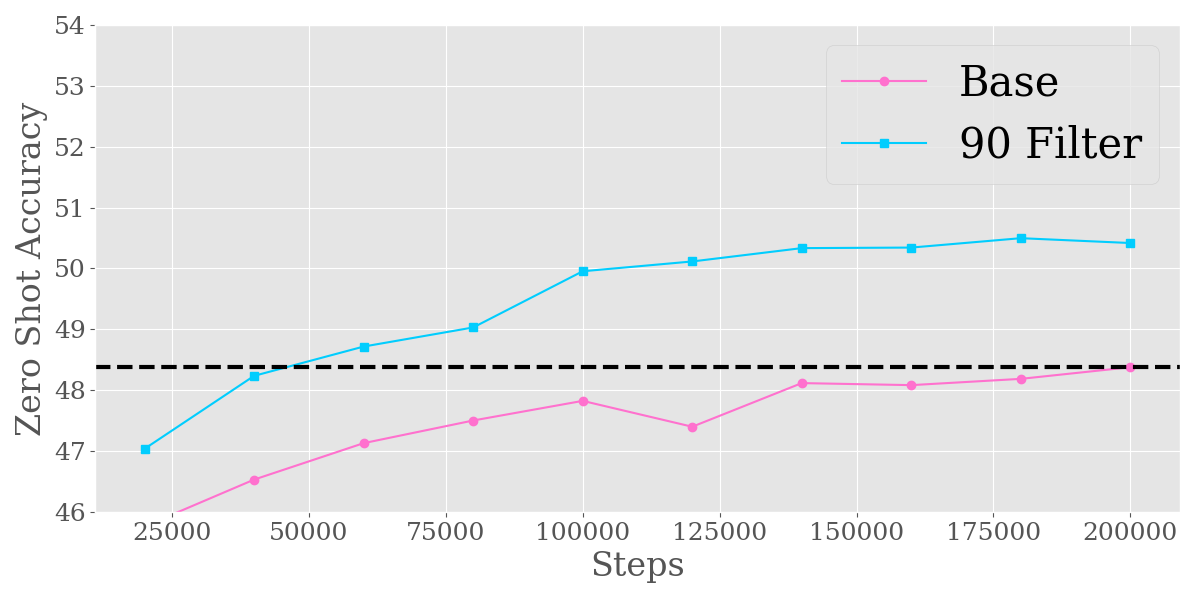}
        \caption{Monolingual FR}
        \label{fig:mono_fr_scaling}
    \end{subfigure}
    \hfill
    \begin{subfigure}{0.3\textwidth}
        \centering
        \includegraphics[width=\textwidth]{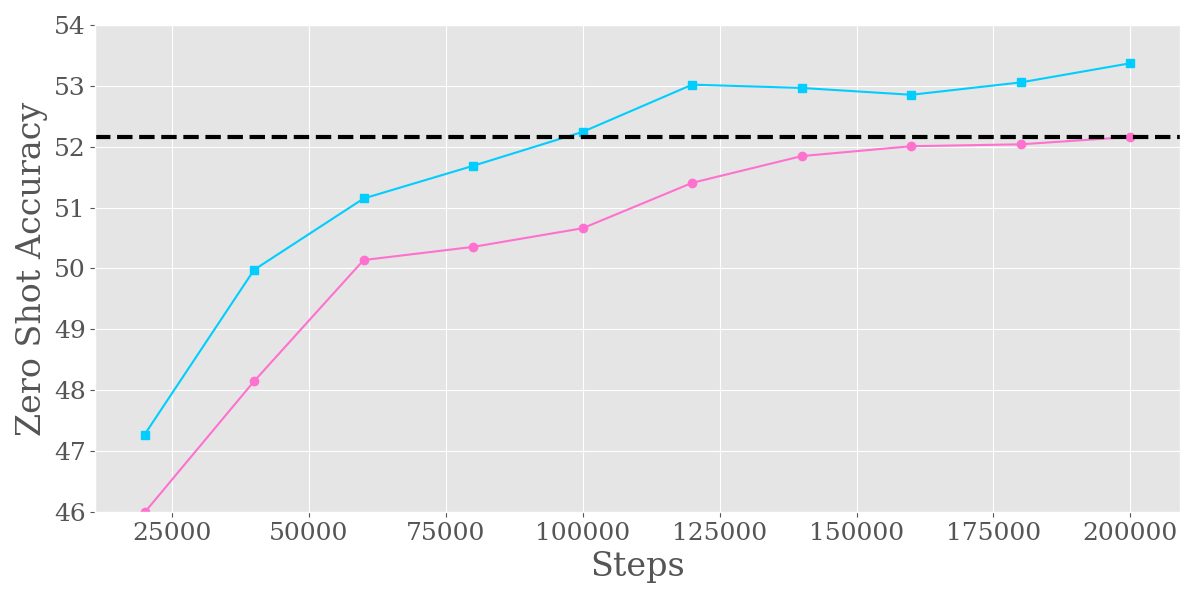}
        \caption{Bilingual FR}
        \label{fig:bi_fr_scaling}
    \end{subfigure}
    \hfill
    \begin{subfigure}{0.3\textwidth}
        \centering
        \includegraphics[width=\textwidth]{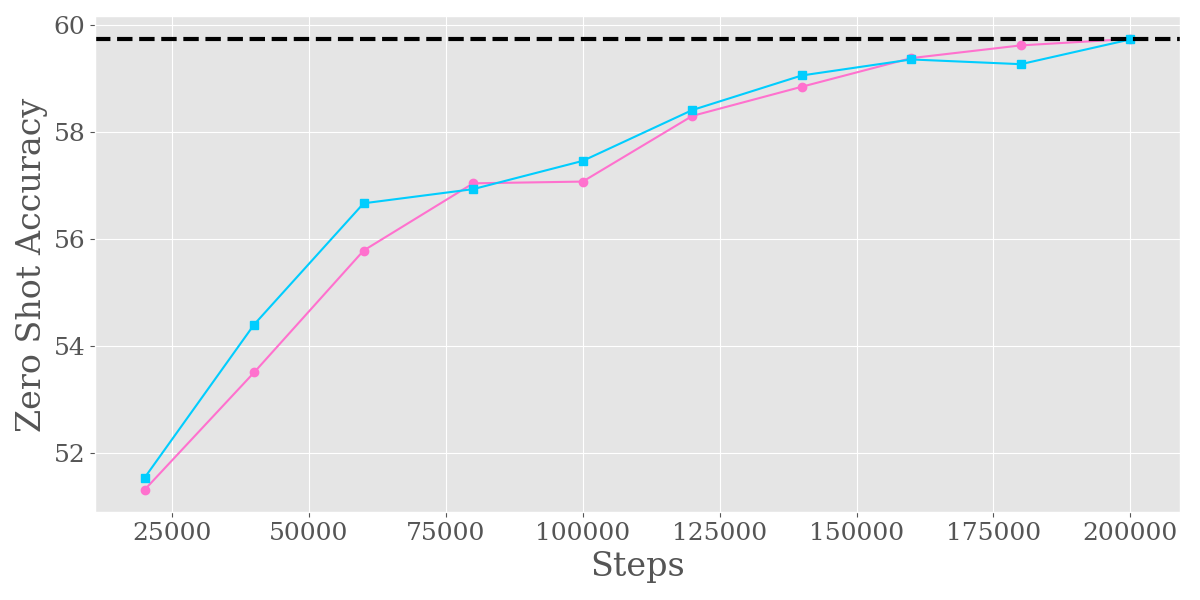}
        \caption{Bilingual EN}
        \label{fig:bi_en_scaling}
    \end{subfigure}
       
    \caption{Performance at intermediate checkpoints during training for 1.3B models for Core EN and FR benchmarks.}
    \label{fig:data_scaling}
\end{figure*}

\paragraph{Findings:} As we increase data quality, both the EN and FR performance increase.  90\% filtering achieves the strongest performance. We note that this is true even over higher filtering where we did train a model at 95\% filtering for RedPajama2, but observed over-filtering on the data as training for the same number of steps requires repeating the data which leads to performance leveling out.  

We further note that while French evaluations improve consistently over base RedPajama2, and are close to TransWebEDU FR, the English performance is worse and consistent with the base comparison from 60\% indicating that the FineWebEDU corpus is still  higher quality as it has some additional filtering over RedPajama2.  For evaluations filtering FineWeb2, we note that performance in English is within 1\%, and performance in French is better than using TransWebEDU.  We conclude that improving data quality using our filtering mechanism leads to performance improvements also in English over low quality data.

\subsection{Comparison with Bi- and Multilingual Models}
\label{sec:public_comparison}
\paragraph{Methodology} We show that our data selection process in Section~\ref{sec:dq} can be used to select high quality data consistent with other public bilingual and multilingual models. We study performance for 1.3B parameter models and similar sizes across a range of tasks in English and French as (i) there are a number of available evaluations that are both translated and native, and (ii) there are other bilingual or multilingual models in French for comparison. All of our models are trained with FineWebEDU English as the English data source for 200K steps, which is up to 15$\times$ fewer than other models.  Additional details on the evaluation sets are provided in Appendix~\ref{sec:appendix_datasets}. For model comparisons, we include strong bilingual models like CroissantLLM \cite{faysse2024croissantllm}, models trained on Indo-European languages like TransWebEDU EN-FR, TransWebLLM \cite{wang2025multilingual}, and EuroLLM \cite{martins2024eurollm}, and multilingual models like Bloom \cite{le2023bloom}, and Qwen2.5 \cite{yang2024qwen2}, all of which achieve strong results on multilingual benchmarks. 

\paragraph{Findings: } Our filtering leads to better zero-shot performance over public bilingual models such as Croissant LLM (1.7\%), and competitive performance (up to 4\% increase) to multilingual models trained for much longer highlighting the benefits of training a bilingual model on high quality data. Our models attain better performance in French than all models except for Qwen2.5 1.5B\footnote{Note that Qwen2.5 is shown to be one of the best performing multilingual models over other open-weight models \url{https://huggingface.co/spaces/HuggingFaceFW/blogpost-fine-tasks}.} which has overall $0.4\%$ improvement while being trained for much longer and on combinations of different data.  However, on English data, the Qwen and EuroLLM performance exceed our models.  

\subsection{Data Scaling}
\label{sec:scaling}

\paragraph{Methodology:} Our main results train models for 200K steps amounting to 200B tokens as  Section~\ref{sec:data_quality} shows benefits from both training for longer and on higher quality toward diminishing bilingual model gap in performance. This amount, although below proprietary models at similar scales such as \cite{touvron2023llama,yang2024qwen2}, is above the recommended training data size according to Chinchilla~\cite{rae2021scaling} ($\sim7$x chinchilla).  Li et al. \cite{li2024datacomp} filter to a small amount of data from a much larger pool (although they filter at the same 10\% rate) and train only at 1-2$\times$ Chinchilla scale thus having a much smaller ratio of tokens used. We now show that performance improvements hold at intermediate training steps.

\paragraph{Findings:}  We show results in Figure~\ref{fig:data_scaling} for monolingual and bilingual models. Both monolingual and bilingual models evaluated on Core French benchmarks show consistent performance improvements at all stages of training.  For  monolingual models,  performance from filtering leads to around $5\times$ efficiency as the model attains the same performance at only 40K steps.  For bilingual models, we note that the higher quality English data reduces that gap consistent with findings on better auxiliary data from \cite{seto2024training}, however we still observe around $2\times$ speedup in training.  Finally, for English evaluations, we observe improvements only early in training consistent with bilingual gaps earlier.  The gap in English performance diminishes  after around 60K steps ($\sim2$x Chinchilla).

\section{Conclusion}
\label{sec:conclusion}

Training a multilingual language model that performs as well as monolingual models is important for building language models that can work for everyone, and facilitate compute efficiency in memory constrained settings where keeping many monolingual models may be infeasible. However it is also more challenging as it necessitates learning multiple distributions of data.  This work provides a simple recipe for selecting high quality data, and demonstrates capability of selecting high quality data in other languages with only high quality English data.  Selecting high quality data with our recipe reduces gaps between monolingual and bilingual models to less than 1\%, and improves monolingual performance. Our work takes a step towards pretraining language models in languages with limited high quality data, and can help more research into closing the gap between multilingual and English-centric language models.

\section{Limitations}
This section lists limitations of our work.

\paragraph{Evaluation data.}
Our evaluations languages other than English rely on translated evaluation sets. Not only does this introduce potential translation mistakes (for example for math or certain scientific terms), the resulting evaluation set also contains cultural biases as has been noted in datasets such as MMLU \cite{singh2024global}. As a result, certain aspects of the evaluation may lead to improved performance when using English auxiliary or translated data. Additionally, translated data often exhibits a distribution different from that of real data in the target languages. We focus on French as there are many native language benchmarks for which models perform well.

\paragraph{Languages included.}
The focus of this work is on training bilingual language models.  We note that there are several languages for which training a bilingual or multilingual language model is now practical given the size and available training data.  However, our goal is in building high quality datasets and showing gaps in performance from lack of data quality control which require filtering from a large pool of data.  Training even a 1.3B parameter model at our scale requires 200B+ tokens of data and filtering down to 10\% of the data leaves only languages with over 2T tokens (for one repetition training), for which there are few.  As we are constrained by having a large pool of tokens with relatively little filtering (Redpajama2), a more highly curated set of tokens (FineWeb2), and native evaluations, for our ablations and studies, this left only a few languages: French, German, and Chinese.  We study French in the main text as it satisfies all conditions, and is relatively close to English indicating potential transferability as shown in \cite{seto2024training}.  We further note that another constraint is the number of languages in our SentenceBert embeddings.  The multilingual SentenceBert model used in this work supports 50+ languages\footnote{A full list of language codes: ar, bg, ca, cs, da, de, el, en, es, et, fa, fi, fr, fr-ca, gl, gu, he, hi, hr, hu, hy, id, it, ja, ka, ko, ku, lt, lv, mk, mn, mr, ms, my, nb, nl, pl, pt, pt-br, ro, ru, sk, sl, sq, sr, sv, th, tr, uk, ur, vi, zh-cn, zh-tw.}. While this already covers many more languages than high performing language models in those languages, there are methods for adding languages to a multilingual embedding via knowledge distillation \cite{reimers2020making}.

\bibliography{main}

\clearpage 
\appendix
\section{Hyperparameters and Additional Training Details}
\label{sec:hyperparams}

The small model is a 350M non-embedding parameter model consisting of 24 layers, 16 attention heads, and a hidden dimension size of 1024. The 1.3B non-embedding parameter model consists of 24 layers, 16 attention heads, and a hidden dimension size of 2048. Both models have a maximum sequence length of 1024.  The 2.7B parameter model consists of 32 layers with 2560 hidden dimension and 32 attention heads.

The baseline models are trained using NVIDIA’s Megatron-LM\footnote{\url{https://github.com/NVIDIA/Megatron-LM}} repository for pretraining language models. All models are trained for a total of 200K steps with a batch size of 1024. The 2.7B models are trained with context of 2048 and other models are trained with 1024.

Models are trained using a maximum learning rate of $0.0003$ for the 350M  model, $0.0002$ for the 1.3B model, and $0.00016$ for the 2.7B models with a minimum learning rate of $0.00001$ with a cosine learning rate scheduler and warmup for $1\%$ of the total steps. For regularization, we use a weight decay of $0.01$, along with a gradient clipping norm of $1.0$. Models are trained with the Adam optimizer using $\beta_1=0.9$ and $\beta_2=0.999$.

The total training time for 1.3B models on roughly 200B tokens is around 2000 GPUh on Nvidia H100 GPUs.  For 350M models, the total training time is around 1200 hours. For a 2.7B model trained on roughly 400B tokens, the total time is around 9000 GPUh on Nvidia H100.

\section{Dataset Details}
\label{sec:appendix_datasets}

\subsection{Training Datasets}
 We consider several datasets in this work and primarily focus on FineWeb2, and RedPajamav2 for pretraining.  We choose these datasets as there exist a sufficiently large amount of data in multiple languages.  For experiments filtering high quality data, we focus on Redpajamav2 as there are up to 3T tokens of data in these datasets compared with mC4 ($\sim 300$B), and FineWeb2 ($\sim206$B words) for French, and the data is native (non-translated).  We also experiment with TransWebEDU \cite{wang2025multilingual} for comparison to training on translated high quality data, and with filtering from FineWeb2, an already filtered but smaller pool of data.  We primarily focus on English-French bilingual pretraining in this section as we have both larger amounts of data for pretraining in publicly available corpora such as RedPajama2 and FineWeb2, have native evaluation sets, and the language is relatively close to English.  We additionally present results with the high quality filter in German and Chinese in Section~\ref{sec:german_chinese_filter_eval}. We choose German as there is also a large amount of high quality data, its closeness to English, and it is commonly used in other works \cite{seto2024training}. However, we note that there is only a small amount of native evaluation data such as Include \cite{romanou2024include} and Kaleidoscope \cite{salazar2025kaleidoscope} for which there are only a few hundred samples. We also evaluate on Chinese to test a language further from English with a large amount of publicly available data, and native evaluations.  We provide a brief description of each dataset below as well as the token counts for the approximate number of tokens used used for each dataset in training\footnote{Note that for training with the Megatron library, we tokenize batches of parquet or jsonl files (referred to as a dataset in Megatron-LM), and use each dataset with equal weight. This means that if some files or documents have fewer tokens, they might repeat at a higher rate than other sets of files. While we do see more repetitions for a few subsets, this is relatively small for overall training, and for training, we still repeat data for only a few epochs less than would incur a gap in performance to single epoch training following \cite{muennighoff2024scaling}.}.

\begin{itemize}[leftmargin=*]
    \item \textbf{mC4}: We use the multilingual Colossal Clean Crawled Corpus (mC4), a curated text dataset comprising over 6.3T tokens for experiments in Section~\ref{sec:data_quality}. This corpus is derived from CommonCrawl and used for pretraining numerous language models~\cite{brown2020language,raffel2020exploring,touvron2023llama}. The dataset is chosen as a low quality dataset as it is relatively little filtering.  For our experiments we use the first $\sim520$ files for translation and otherwise train on one epoch or two epoch of data from this subset \cite{xue2020mt5}.
    \item \textbf{FineWebEDU}: A subset of the FineWeb dataset which is filtered according to a classifier trained on annotations for educational quality from Llama-3 70B model~\cite{penedo2024fineweb}. We use the subset known as TransWebEDU, which is a subset of around 75B tokens used in \cite{wang2025multilingual}.  We also use the machine translated German version and translate using a proprietary translation system into Chinese in Section~\ref{sec:data_quality}. We use all files from this dataset given the already smaller size.
    \item \textbf{ChineseFineWeb-EDU}: An educational corpus in Chinese consisting of roughly 400B tokens of data.  Although it shares a similar name, the ChineseFineWeb-EDU does not share data from FineWebEDU and is collected from different sources.  We use the first 600 files in total for our experiments \cite{yu2025opencsgchinesecorpusseries}.
     \item \textbf{RedPajama2}: A pretraining corpus with light filtering consisting of 30T tokens: 20T tokens of English text, and $\sim3$T for German and French.  We focus on the French and German portions of the dataset only. We randomly shuffle all subsets of the data and train using a random shuffled subset of both the head and middle portions~\cite{weber2024redpajama}. 
     \item \textbf{FineWeb2}:  Data sourced in a similar way as FineWeb but for many languages.  We use French, German, and Chinese subsets.  The French data has 113 parquet files, German has 122, and Chinese has 185 parquet files. Given the size of the datasets, data is repeated for multiple epochs though still under 10 epochs to not yield worse performance than training on new data following \cite{muennighoff2024scaling}.
\end{itemize}

\begin{table}[ht]
\centering
\begin{tabular}{lcccc}
\toprule
\textbf{Dataset} & \multicolumn{4}{c}{\textbf{Tokens (B)}} \\
& \textbf{EN} & \textbf{FR} & \textbf{DE} & \textbf{ZH} \\
\midrule
mC4 EN  &       125      &    76        &  75           &        --     \\
RPJ2 (base)     &   --       &      310    & 297          & --          \\
RPJ2 (90\%)     &   --       &      260    & 248           & --          \\
FineWeb2   &   --        &    270      &  260         &   282      \\
FineWeb2 (90\%)   &   --        &    34       &     28      &   30      \\
TransWebEDU &    54      &    62       &  55         &      45    \\
ChineseFineWeb &  192     &     --      &    --       &     195     \\
\bottomrule
\end{tabular}
\caption{Upper bound on the approximate number of tokens by language used in training in this work for  training datasets used in this work.}
\end{table}

\subsection{SentenceBert Filter Scores}

We report the SentenceBert filter scores corresponding to different percentiles for all datasets we filter.  Filter scores are primarily estimated using only the first file, however we also compare this with filter scores from 100 randomly selected files and find that they are similar.  We report the scores used in Tabl~\ref{tab:filter_scores}.  Note that for some datasets, we only conduct experiments using the 90th percentile.  

\begin{table*}
    \centering
\begin{tabular}{lccccc}
    \toprule 
    \textbf{Percentile} & \textbf{RPJ2 FR} & \textbf{RPJ2 DE} & \textbf{FW2 FR} & \textbf{FW2 DE} &  \textbf{FW2 ZH} \\
    \midrule
    95 & 0.4014 &  & & & \\
    90 & 0.2170 & 0.1654 & 0.2920 & 0.2651 & 0.2751 \\
    70 & 0.0610 & 0.033931 & 0.0884 & 0.0614 & 0.0722 \\
    60 & 0.0361 & 0.019172 & 0.0546 & 0.0355 & 0.0437 \\
    40 & 0.0133 & 0.006802 & 0.0237 & 0.0130 & 0.0170 \\
    30 & 0.0077 & 0.003946 & 0.0140 & 0.0077 & 0.0107 \\
    10 & 0.0017 & 0.000884 & 0.0029 & 0.0019 & 0.0028 \\
    \bottomrule
\end{tabular}
\caption{Filter percentile scores for different datasets.}
\label{tab:filter_scores}
\end{table*}

\subsection{Zero Shot Evaluations}
\subsubsection{Core Benchmarks}
\begin{itemize}[leftmargin=*]
    \item \textbf{SciQ [Core]}: A dataset of science exam questions for evaluating the ability of NLP models in understanding and reasoning within the science domain~\cite{welbl2017crowdsourcing}.
    \item \textbf{ARC Challenge (ARC-C) [Core]}:Part of the AI2 Reasoning Challenge (ARC)~\cite{clark2018think}, containing science exam questions from grades 3 to 9. The ARC Challenge set includes more difficult questions that necessitate higher-order reasoning.
    \item \textbf{ARC Easy (ARC-E) [Core]}: The Easy set of the AI2 Reasoning Challenge~\cite{clark2018think} features questions from the same source as ARC-C but are considered less challenging.
    \item \textbf{Winogrande (WG) [Core]}: This dataset challenges models on common sense reasoning in a language context, focusing on pronoun disambiguation tasks~\cite{sakaguchi2021winogrande}.
    \item \textbf{PIQA [Core]}: Physical Interaction Question Answering tests the understanding of everyday physical processes~\cite{bisk2020piqa}.
\item \textbf{HellaSwag (HS) [Core]}: Evaluates a model's ability to complete scenarios in a contextually and logically coherent manner~\cite{zellers2019hellaswag}.
\end{itemize}
We use the same translations from (Anonymous, 2024).  For our evaluations, we use the lm-eval-harness repository\footnote{\url{https://github.com/EleutherAI/lm-evaluation-harness}} for zero-shot accuracy on QA tasks.

\subsubsection{Other Evaluation Datasets}
\begin{itemize}[leftmargin=*]
    \item \textbf{MMLU}: Multi-domain question answering, MMLU assesses the model's expertise over a wide range of specialized subjects, from professional domains to academia~\cite{hendrycks2020measuring}.  We use the human translated versions available from GlobalMMLU \cite{singh2024global}.
    \item \textbf{FrenchBench-MC}: Collection of four evaluations including translated versions of ARC-challenge, HellaSwag, grammar, and vocab \cite{faysse2024croissantllm}.
    \item \textbf{Regional}: Evaluation on both the Include \cite{romanou2024include} and Kaleidoscope \cite{salazar2025kaleidoscope} benchmarks. For Kaleidoscope, we use only the portion that does not require image modality. As both evaluation sets are small and test regional knowledge, we group both and average the accuracy.
    \item \textbf{NLI}: We report accuracy over French topic-based NLI \cite{faysse2024croissantllm}, and XNLI \cite{conneau2018xnli} translated into French.  
\end{itemize}

\subsection{Licenses and Attributions}
The training datasets are supported by public licenses including ODC and Apache license.  The pre-trained models including Mistral (for translation), SentenceBert, and OH FastText classifiers are also supported by Apache and MIT licenses. The translated data for Section~\ref{sec:data_quality} uses a proprietary translation model following \cite{seto2024training}.

All models and datasets are collected from Huggingface via the datasets library, and all models are evaluated using the lm-eval-harness library from EleutherAI \cite{eval-harness}, which uses an MIT license.

We use the Megatron codebase under the Nvidia license for pre-training.

\section{Curse of Multilinguality}
Early works found that pretraining language models on a large number of languages leads to a decrease in performance for each language \cite{conneau2019cross,rust2021good,wang2020negative,chang2024multilinguality}.    Several works have investigated causes of performance degradation \cite{rust2021good,wang2020negative,chang2024multilinguality}, and methods for addressing this \cite{blevins2024breaking,pfeiffer2022lifting}.  Our work focuses on bilingual language model performance degradation, which limits to a degree the impact of many languages and focuses instead on  data size, quality, and training time.  While our work can help shed light on factors impacting multilingual model training, our focus is on mitigating performance gaps and the reason for these gaps.

\section{SBERT Classifier Embeddings for Classification}
\label{sec:sbert_classifier_details}
Prior works have used SBERT for training a linear classifier \cite{minaee2021deep,albalak2024survey,grangier2024task}, but only in the same language and do not its impact in multilingual LM learning.  We show that training a quality classifier with only English data is feasible.

\begin{figure}
    \centering
    \includegraphics[width=\linewidth]{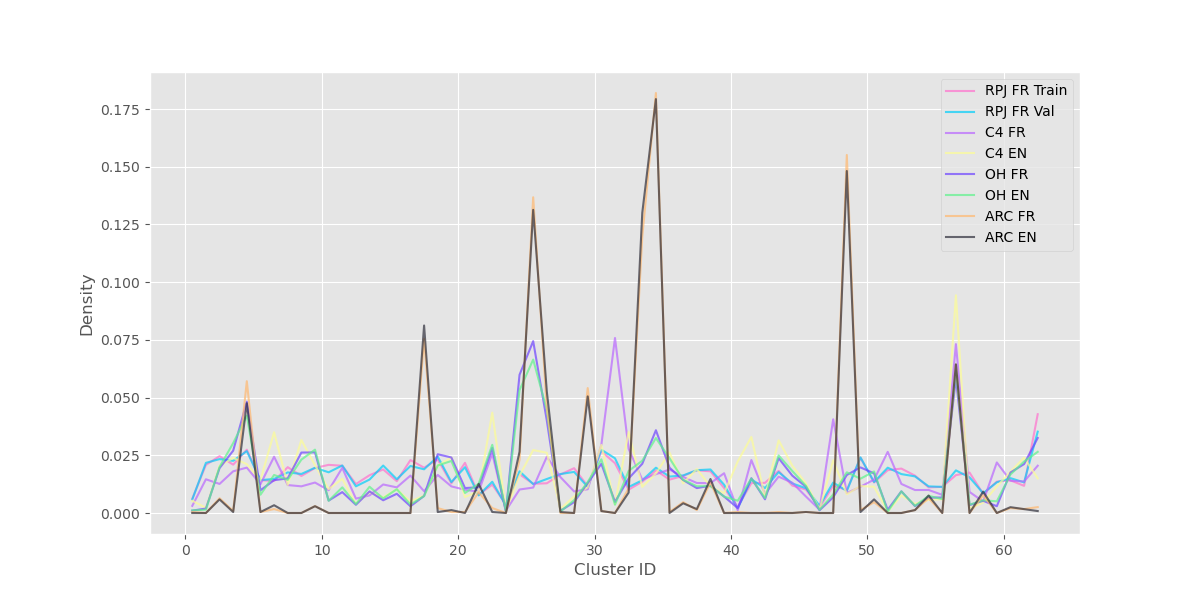}
    \caption{Cluster histograms for distribution of different datasets over RedPajama2.  Both English and French versions of the data have similar distributions}
    \label{fig:hist_data}
\end{figure}

We train a K-means clustering with 64 balanced clusters over the embeddings of 10 files of RedPajamav2 French data to examine the distribution of different datasets following \cite{grangier2024specialized}.  We then label sets of data in both English and French including C4 \cite{xue2020mt5}, DCLM classifier training data \cite{li2024datacomp}, and ARC Easy\cite{clark2018think}.  Figure~\ref{fig:hist_data} shows that both French and English data follow similar histograms indicating that data lie close in the same clusters and can be interchanged when filtering data.  As a result we will be able to select the same distribution of data for training models.

\section{German and English Data Quality Experiments}
\label{sec:heatmap_german}

We start replicating setup in which a bilingual language model is trained on an equal proportion of data from mC4 in  German (DE) and English (EN) totaling 100K steps each, following the setup for French and English presented in Section~\ref{sec:data_quality}.

\begin{table}[ht]
\centering
\resizebox{\columnwidth}{!}{
    \begin{tabular}{lcccc}
        \toprule
        \textbf{Model} & \textbf{Core EN} & \textbf{Core DE} &  \textbf{MMLU EN} & \textbf{MMLU DE} \\
        \midrule
        EN & \textbf{56.2} & 40.9 & \textbf{29.8} & 26.4 \\
        DE & 46.5 & 48.9 & 26.8 & 27.4 \\
        BI & 53.3 & \textbf{49.7} & 28.9 & \textbf{28.0} \\
        \bottomrule
    \end{tabular}
}
\caption{Zero shot accuracy for general understanding and specialized knowledge tasks for monolingual English (EN), German (DE), and bilingual (BI) models.}
\label{tab:mc4_de_results}
\end{table}

Table~\ref{tab:mc4_de_results} shows performance on MMLU and Core benchmarks.  Our findings match those in prior works where we see a 3\% drop in English and an increase in German of 1\% compared to the bilingual model. 

Next, we control the data quality and languages in the model.  We follow the same setup as in French and vary both the data quality and the language with mC4 and FineWebEDU translations using the same translation system for mC4 and TransWebEDU translations for German.

\begin{figure*}[ht]
    \centering
       \begin{subfigure}{0.23\textwidth}
        \centering
        \includegraphics[width=\textwidth]{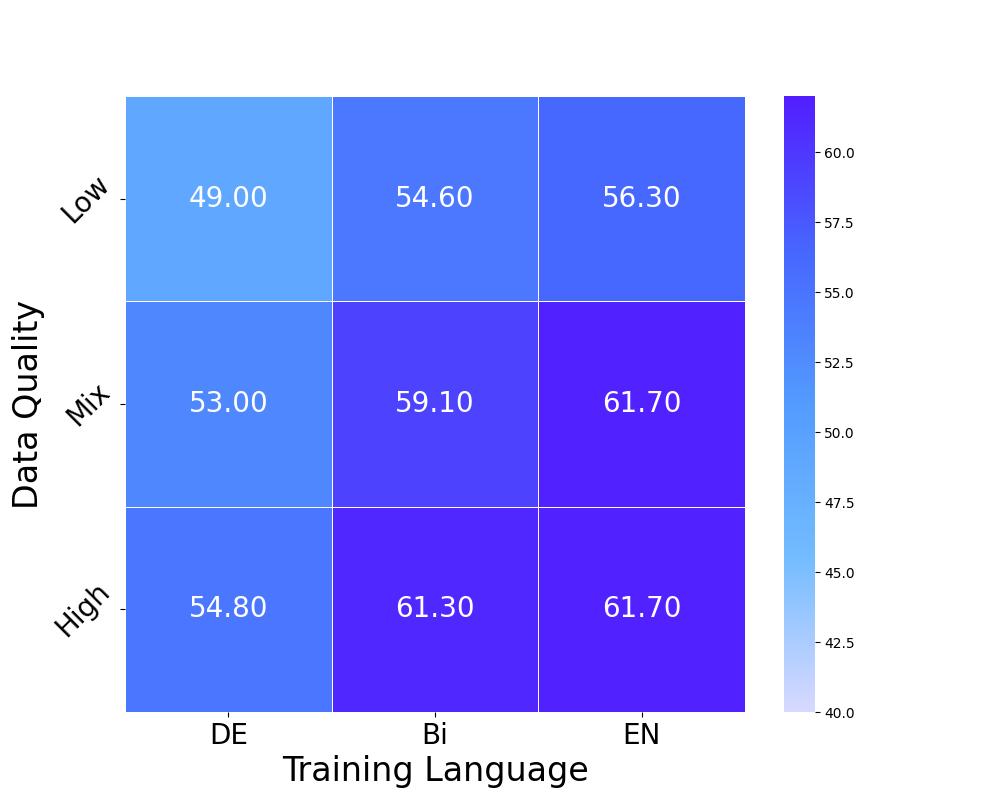}
        \caption{English Core}
        \label{fig:en_core_200}
    \end{subfigure}
    \hfill
       \begin{subfigure}{0.23\textwidth}
        \centering
        \includegraphics[width=\textwidth]{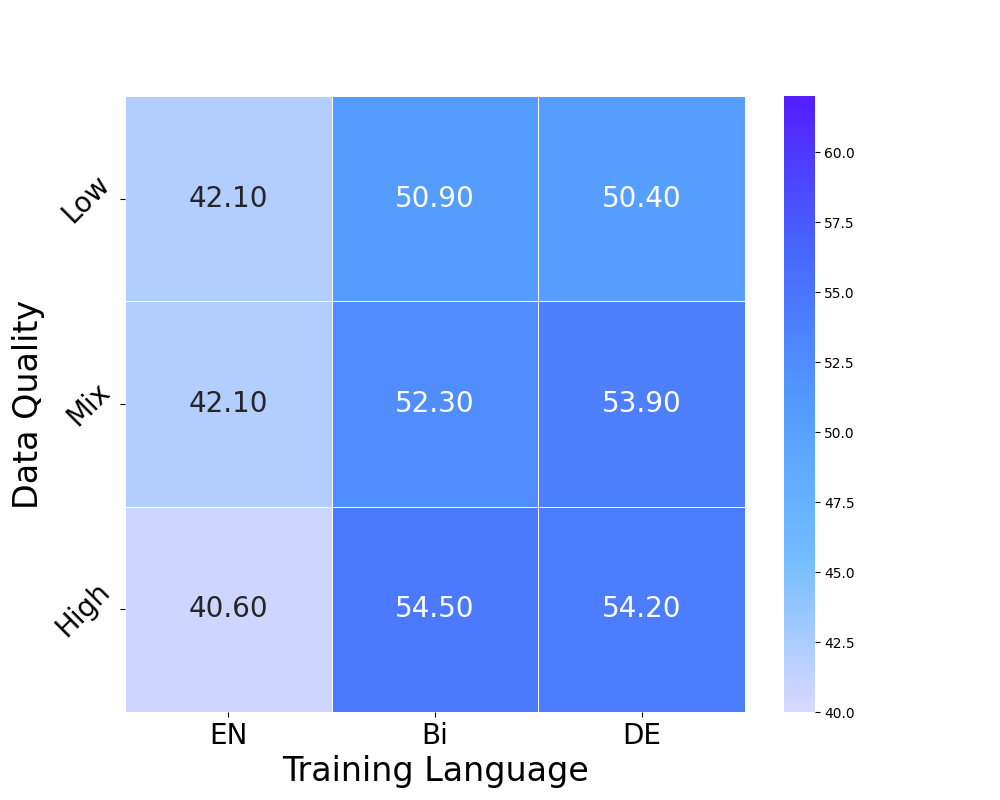}
        \caption{German Core}
        \label{fig:de_core_200}
    \end{subfigure}
    \hfill
       \begin{subfigure}{0.23\textwidth}
        \centering
        \includegraphics[width=\textwidth]{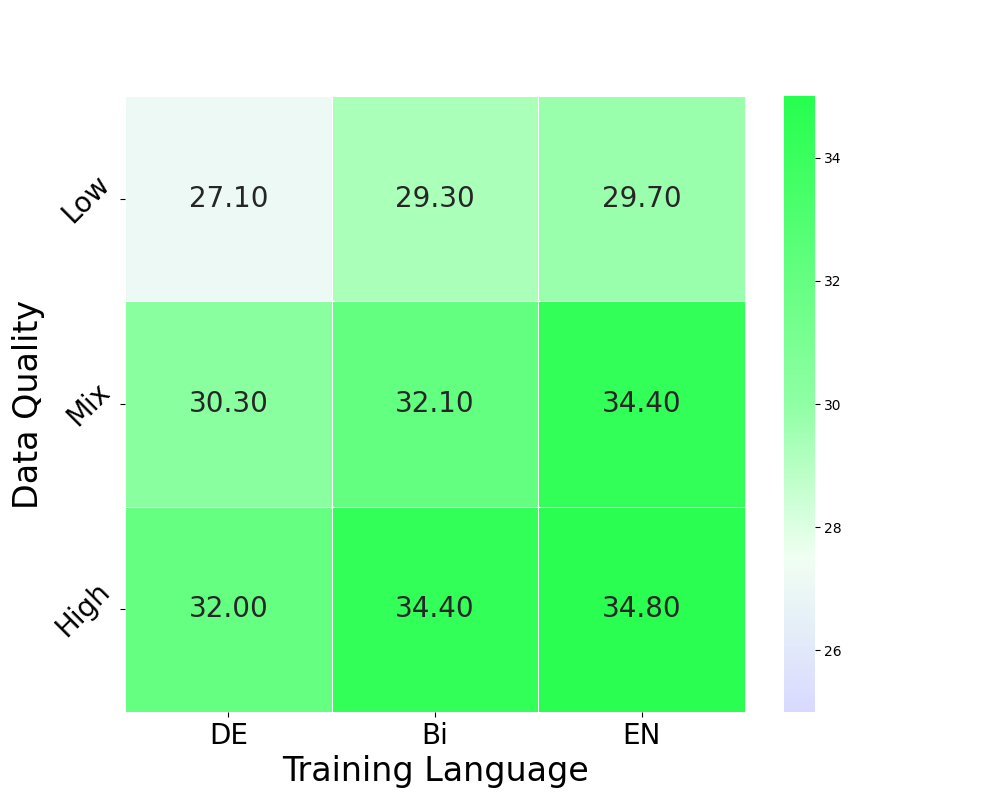}
        \caption{English MMLU}
        \label{fig:en_mmlu_200}
    \end{subfigure}
    \hfill
       \begin{subfigure}{0.23\textwidth}
        \centering
        \includegraphics[width=\textwidth]{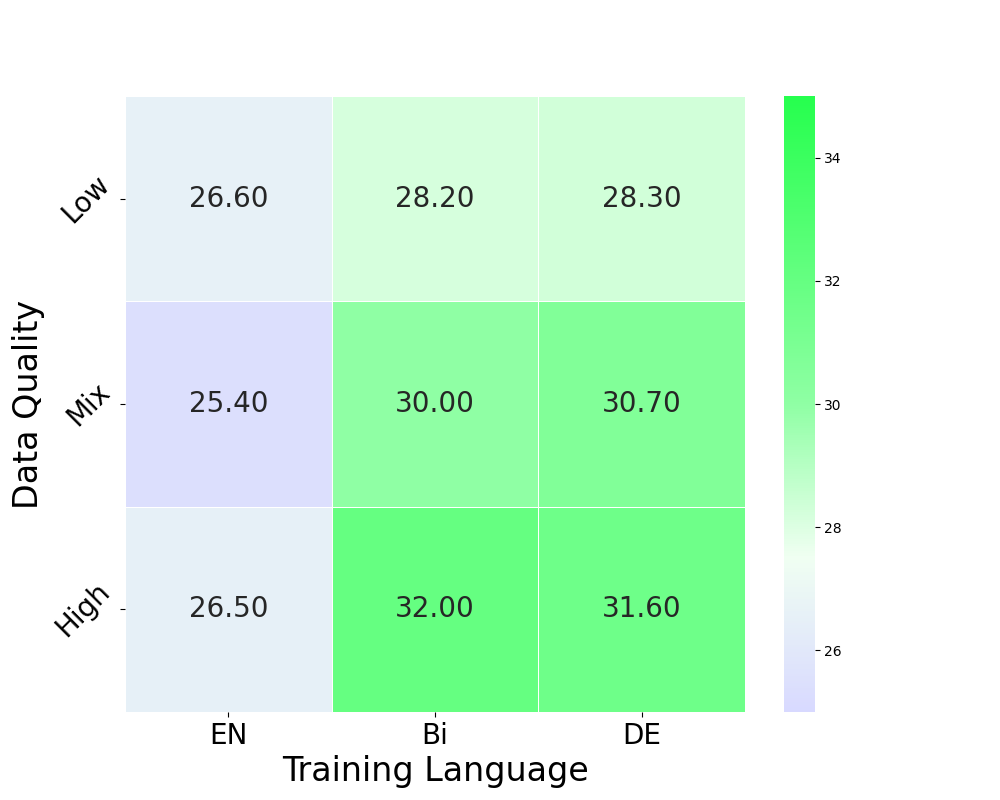}
        \caption{German MMLU}
        \label{fig:de_mmlu_200}
    \end{subfigure}
       
    \caption{Performance with varying data quality and language.  Models are trained on combinations of mC4 (low) and FineWebEDU (high) in native English (EN) and translated to German (DE). Models are trained for 200K steps and evaluated on Core (avg over six common-sense reasoning tasks) and MMLU.}
    \label{fig:en_de_heatmaps_200K}
\end{figure*}

Figure~\ref{fig:en_de_heatmaps_200K} shows the performance difference when varying quality (y-axis) and language (x-axis) for two sets of zero-shot evaluations: Core and MMLU.  When examining the plots, we see that the bottom right square corresponding to a monolingual model trained on high quality in the targeted language has the highest performance.  This is closely followed by models which individually vary the language but keep high quality [middle bottom square, e.g., (bi, high)] or mix quality but keep the same language [right middle square, e.g., (EN, mix)]. However, mixed quality and mixed language taken together [middle square, e.g., (bi, mix)] exhibits an average 2.5\% drop in English performance by comparison.

\begin{figure*}[ht]
    \centering
       \begin{subfigure}{0.23\textwidth}
        \centering
        \includegraphics[width=\textwidth]{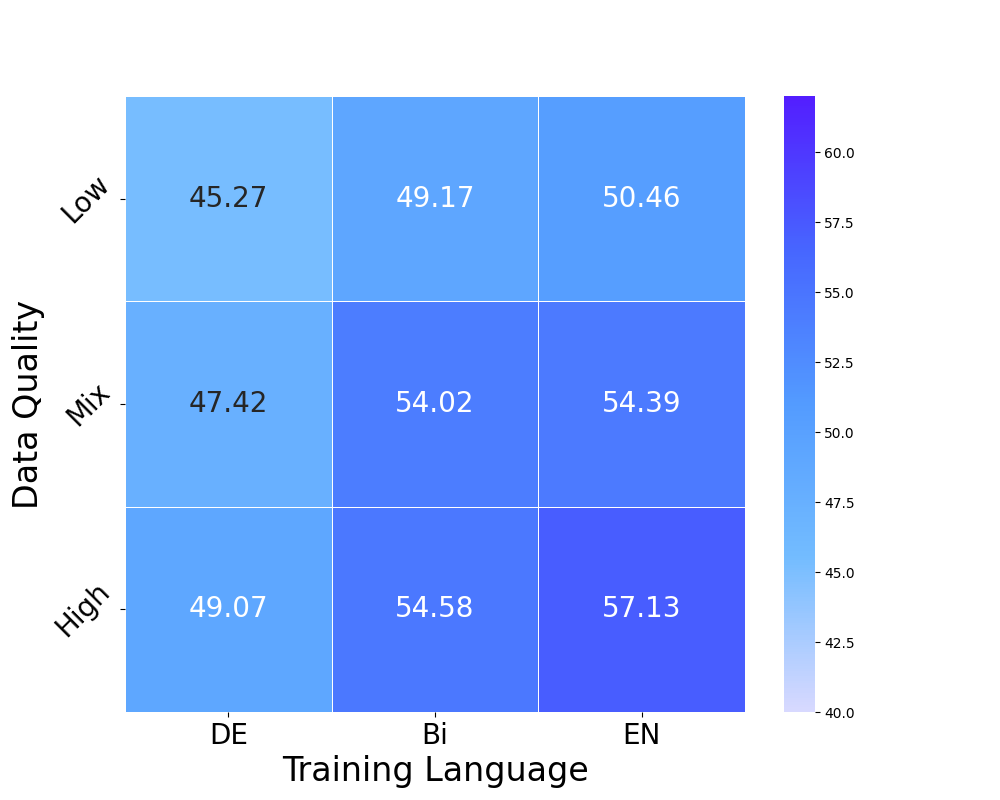}
        \caption{English Core}
        \label{fig:en_core_30}
    \end{subfigure}
    \hfill
       \begin{subfigure}{0.23\textwidth}
        \centering
        \includegraphics[width=\textwidth]{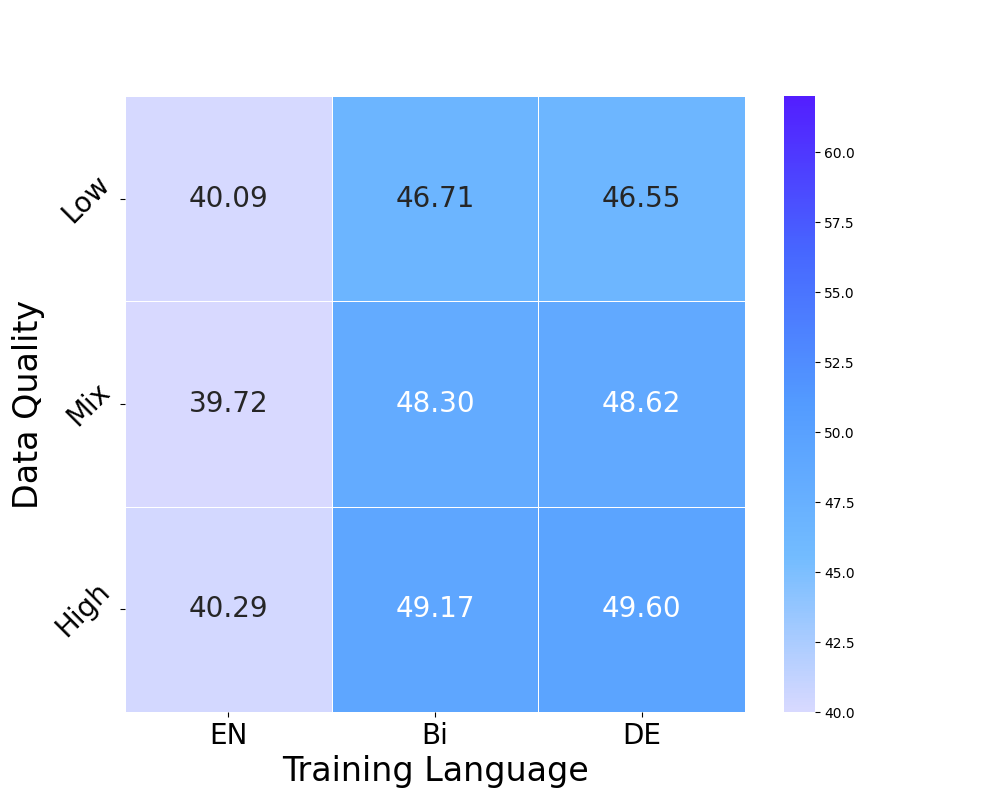}
        \caption{German Core}
        \label{fig:de_core_30}
    \end{subfigure}
    \hfill
       \begin{subfigure}{0.23\textwidth}
        \centering
        \includegraphics[width=\textwidth]{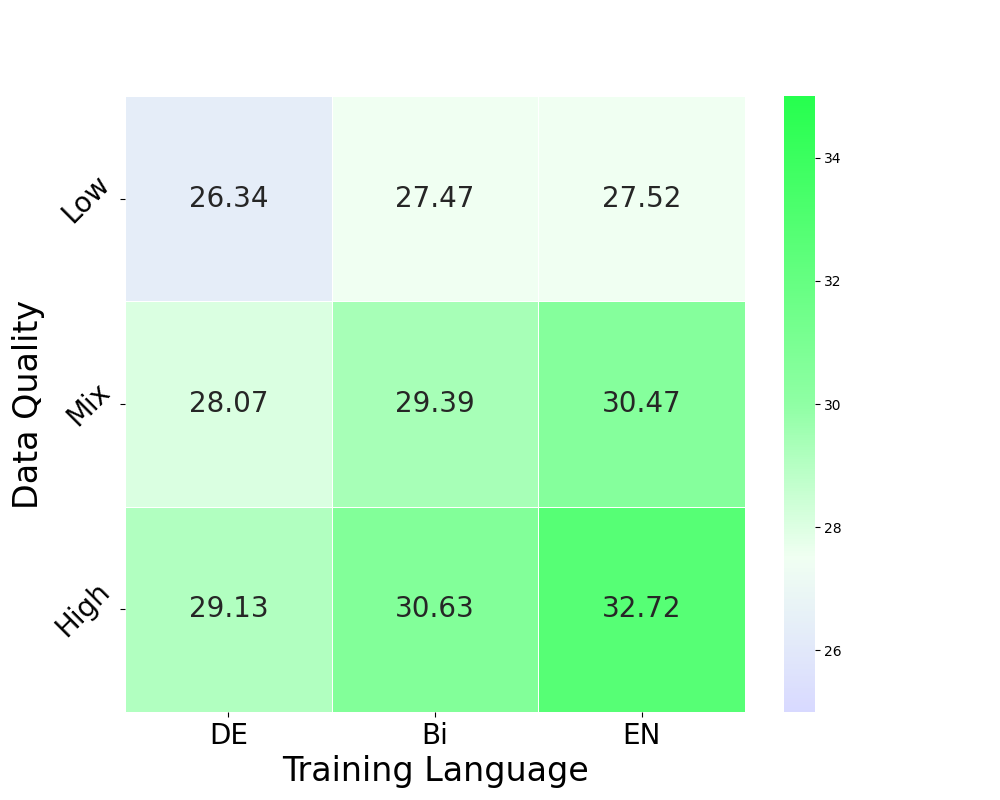}
        \caption{English MMLU}
        \label{fig:en_mmlu_30}
    \end{subfigure}
    \hfill
       \begin{subfigure}{0.23\textwidth}
        \centering
        \includegraphics[width=\textwidth]{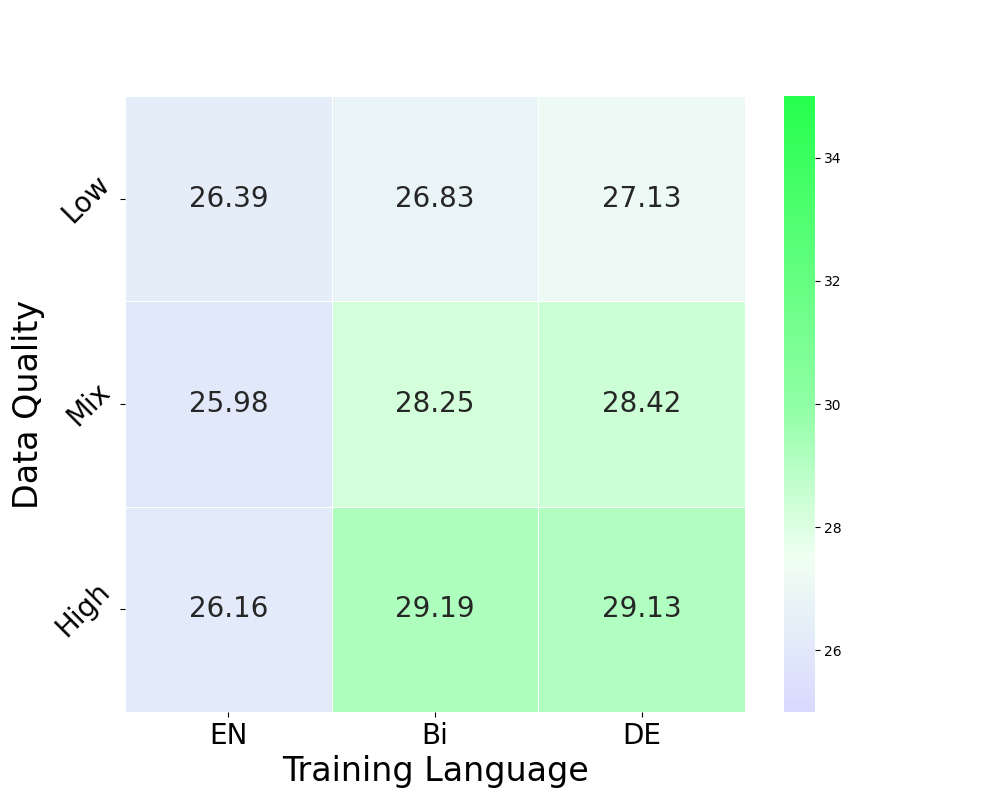}
        \caption{German MMLU}
        \label{fig:de_mmlu_30}
    \end{subfigure}
       
    \caption{Performance with varying data quality and language.  Models are trained on combinations of mC4 (low) and FineWebEDU (high) in native English (EN) and translated to German (DE). Models are trained for 30K steps and evaluated on Core  and MMLU.}
    \label{fig:en_de_heatmaps_30K}
\end{figure*}

Finally, we show results at 30K steps for English and German following the analysis for French.  Results are shown in Figure~\ref{fig:en_de_heatmaps_30K}.  At this scale, we see that both the bilingual high quality (middle bottom) and mixed quality native monolingual (right middle) models have 2.5\% lower performance than the monolingual high quality unlike prior results at 200K steps for English evaluations.  Similarly low quality results (top row) drop below bilingual again indicating data quality has a large role in training.

\section{Additional Filter Results}
\label{sec:additional_filter}

This section presents additional filter percentile results for German and French at larger steps.

\subsection{Filter Percentile Results at 200K Steps}
\begin{figure}[ht]
    \centering
    \includegraphics[width=0.75\linewidth]{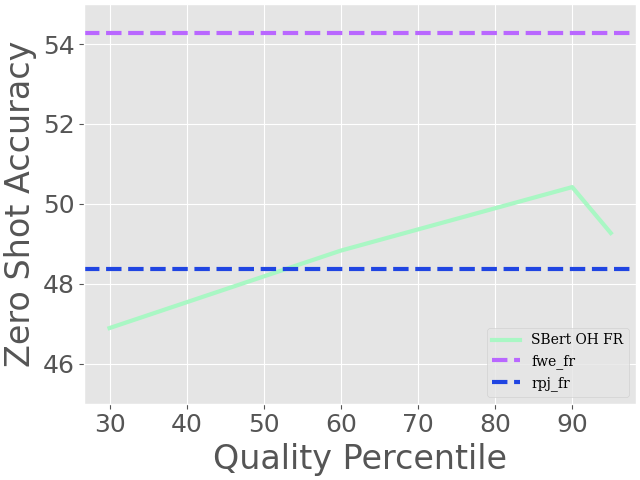}
    \caption{Quality vs. Zero-shot Accuracy on Core tasks for filtered RedPajama2 in French (SBert OH FR) compared with TransWebEDU (fwe\_de) and base RedPajamav2 in French (rpj\_fr) after 200K steps.}
    \label{fig:sbert_filter_fr_200K}
\end{figure}

Figure~\ref{fig:sbert_filter_fr_200K} presents results for different filter percentiles at 200K steps for monolingual French models.  We see that increasing the percentile used in filtering increases performances on benchmarks.  The model performs better than base RedPajamav2 at around 50\% quality filter.  However, the filtered data models perform worse than training for 200K steps on TransWebEDU on translated benchmarks.  Finally, we note that at 95\% quality percentile we observe a plateau in performance where the 90th percentile performs better by $\sim 1\%$. 

\label{sec:filter_200K}

\subsection{FineWeb2 French Filter Percentile Results at 200K Steps}
\label{sec:filter_fw2_200K}

\begin{figure}[ht]
    \centering
    \includegraphics[width=0.75\linewidth]{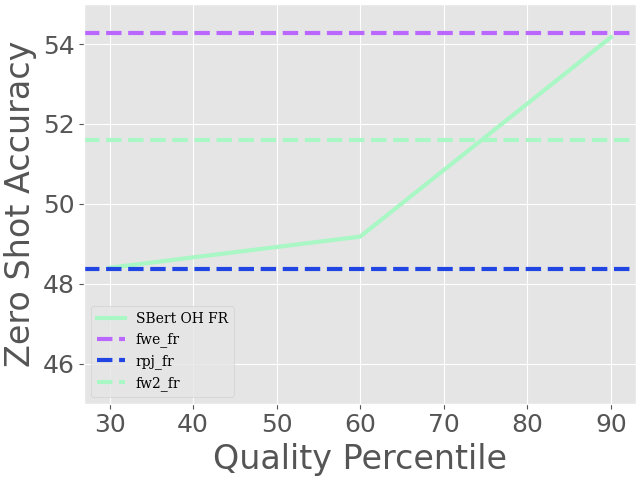}
    \caption{Quality vs. Zero-shot Accuracy on Core tasks for filtered FineWeb2 in French (SBert OH FR) compared with TransWebEDU (fwe\_fr), base FineWeb2 in Frnech (fw2\_fr), and base RedPajamav2 in French (rpj\_fr) after 200K steps.}
    \label{fig:sbert_filter_fw2_fr_200K}
\end{figure}

Figure~\ref{fig:sbert_filter_fw2_fr_200K} presents results for different filter percentiles at 200K steps for monolingual French models.  We see that increasing the percentile used in filtering FineWeb2 increases performance on benchmarks.  The model performs better than base RedPajamav2 at 30\% quality filter, and better than base FineWeb2 at 70\% quality filter percentile.  The filtered data models achieves the same performance a TransWebEDU on translated benchmarks at the 90th percentile.  

\subsection{RedPajamav2 German Filter Percentile Results}
\label{sec:filter_rpjde_30K}

Figure~\ref{fig:sbert_filter_de_30K} presents results for different filter percentiles at 30K steps for monolingual German models. We compare the SBert classifier with training on translated data following the same recipe as for training the original DCLM filter.  We see that increasing the percentile used in filtering increases performances on Core tasks across all filters. However, translating with a  weak translation system appears to plateau performance, with the small CPU translation system with filtering attaining the same performance as the model with no filtering.  Training with a better translations system such as Mistral-7B improves performance to SBert, but requires translating with a more expensive translation system. Second, model trained with SBert filtered data performs better than base RedPajamav2 at around 60\% quality filter.  However, the filtered data models perform worse than training for 30K steps on TransWebEDU (German) on translated benchmarks consistent with our experiments on French.  

\begin{figure}[ht]
    \centering
    \includegraphics[width=0.75\linewidth]{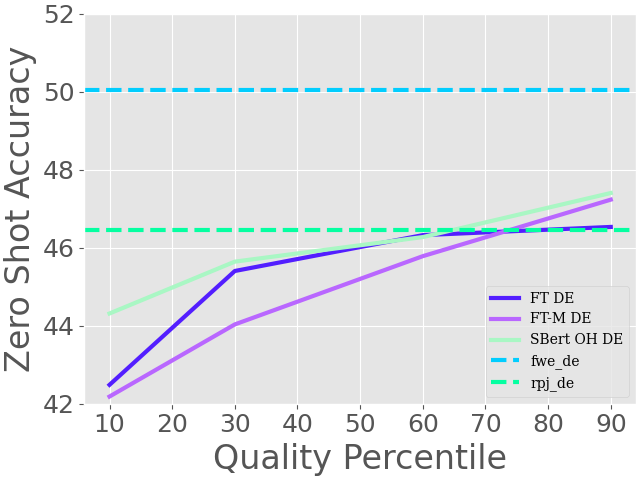}
    \caption{Quality vs. Zero-shot Accuracy on Core tasks for filtered RedPajama2 in German (SBert OH DE) compared with filtering using a FastText classifier trained on translated DCLM classifier training data (FT DE and FT-M DE), TransWebEDU (fwe\_de) and base RedPajamav2 in German (rpj\_de) after 200K steps.}
    \label{fig:sbert_filter_de_30K}
\end{figure}

\section{Continued Pretraining on High Quality Data}
We additionally examine performance when pretraining on the base RedPajamav2 without filtering and subsequently continue pretraining.  We experiment with a 1.3B parameter model and examine continuing pretraining after 150K steps, and after 200K steps.  We report results in Figure~\ref{fig:continue_pretrain}.  Results indicate that after only a few steps (20-30K) we see performance increase consistent with pretraining on filtered data from scratch indicating computational gains if a pretrained model already exists.

\begin{figure}
    \centering
    \includegraphics[width=0.75\linewidth]{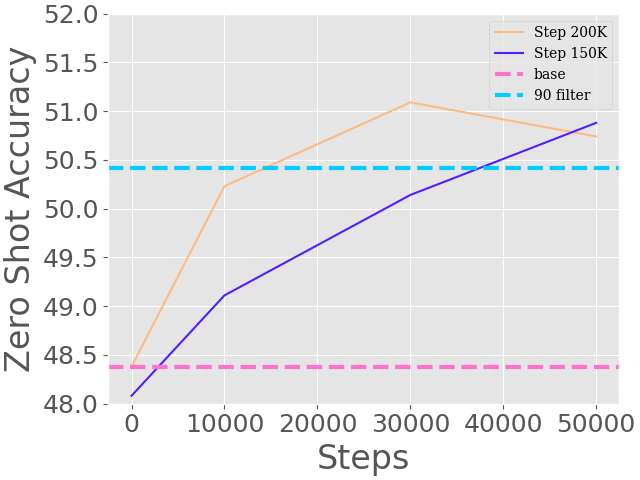}
    \caption{Continued pretraining experiments  for 1.3B model continuing training from 150K steps and 200K steps for a total of 50K steps.}
    \label{fig:continue_pretrain}
\end{figure}

\section{Results with FWE Training Data for Quality Classifier}
\label{sec:fwe_training_data}
\begin{figure}
    \centering
    \includegraphics[width=0.75\linewidth]{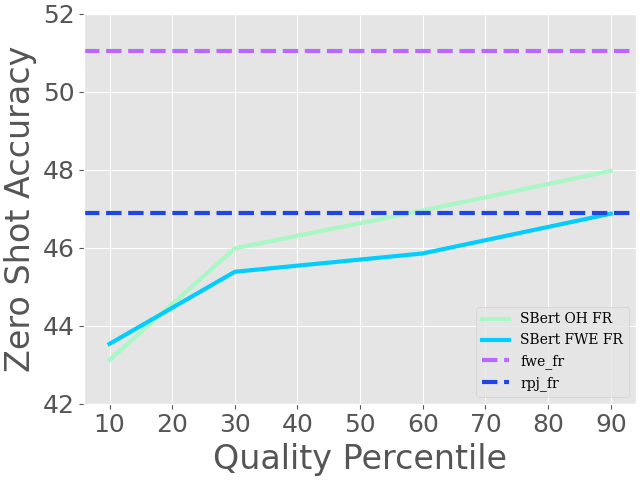}
    \caption{Comparison between training a quality classifier using the DCLM classifier data, and the FineWebEDU data at 30K steps on the Core benchmark datasets for French.}
    \label{fig:fwe_oh}
\end{figure}

Our primary experiments use data from the DCLM classifier training for defining high quality data.  However, there may be several definitions of quality.  We analyze one possible alternative: textbook quality data as defined by FineWebEDU.  We use the same data and annotations used to train the FineWebEDU classifier \cite{penedo2024fineweb}\footnote{The data is available at \url{https://huggingface.co/datasets/HuggingFaceFW/fineweb-edu-llama3-annotations}.}.  We train a binary classifier using the same recipe as prior where the quality label is whether or not the annotation was 2 or above.  Our results are in Figure~\ref{fig:fwe_oh}.  

We find that the DCLM classifier data performs better, while the FineWebEDU data attains the same performance as the base model.  There are a few possibilities that we leave to future work: \begin{enumerate*}[label=(\roman*)]
    \item The original FineWebEDU classifier scores between 0 and 5.  When training a binary classifier, an accuracy of 82\%  is achieved for scoring 3 and above as high quality.  Decreasing to a 2, might make the task harder, resulting in more low quality examples being selected for training.  We chose a classifier score of 2 for high quality as this corresponds to only 30%
    \item Classifying textbook quality data might be a more challenging task when using a universal embedding as the FineWebEDU classifier is a much larger classifier (also an embedding model) than the FastText classifiers.  It's possible that with a more complex classifier, and finetuning performance might be higher.
\end{enumerate*}

\section{Comparison with FW2 HQ}
We additionally compare with the data selection method of \cite{messmer2025enhancing}.  This selection method is similar to the embedding classification approach used in this work, and both build upon the classification method in \cite{grangier2024task}. However \cite{grangier2024task} only study filtering English data and for specific domains. \cite{messmer2025enhancing} applied the filtering to other languages than English from FineWeb2, and although they aim to filter for high quality data in general, they use datasets such as MMLU and Include, making the filtering aimed more at specialization as in \cite{grangier2024task} which also uses MMLU for selection.  They are also primarily focused on monolingual performance, and follow the same regime of using data from the target language for training the classifier, and only test monolingual performance. Collectively we refer to their dataset as \textit{FineWeb2 HQ}.  We train monolingual and bilingual models for the same 200K steps using the dataset made available\footnote{\url{https://huggingface.co/datasets/epfml/FineWeb2-HQ}} for French.  We compare with the same percentile of filtering (90\%) which should yield approximately the same amount of data. We evaluate on the Core benchmarks in Table~\ref{tab:fw2_fw2hq}. 

\begin{table}[ht]
\centering
\small
\begin{tabular}{lcc}
    \hline
    \toprule
    \textbf{Model} & \textbf{Core EN} & \textbf{Core FR} \\
    \midrule
    FW2 & -- & 51.61\\
    FW2 HQ & -- & 52.9 \\
    (Ours) FW2 90\% & -- & \textbf{54.2} \\
    \midrule
    FW2 & 60.26 & 53.65  \\
    FW2 HQ &  \textbf{61.2}&  54.9 \\
    (Ours) FW2 90\% & 61.1 & \textbf{55.5}\\
    \bottomrule
\end{tabular}
\caption{Performance across Core benchmarks comparing models trained with filtering from our approach and \cite{messmer2025enhancing} on FineWeb2.  Top rows represent monolingual performance, and bottom are multilingual with FineWebEDU as the English data.}
\label{tab:fw2_fw2hq}
\end{table}

Our findings show similar performance in English, and that our filtering achieves better performance in French, especially for monolingual. Noting that, we are able to achieve better performance without access to high quality data from other languages for training\footnote{Note that we did not conduct full comparison to other tasks in our list of benchmarks as both MMLU and Regional evaluations have data used for selection}. We hypothesize that this may be because using data from MMLU and Include are very task specific, rather than general high quality data.  Include has specific regional knowledge such as driving exams, which may be filtered for instead, and MMLU is predominantly science knowledge and Western (possibly American) culturally influenced.  Thus, there may be similar issues as we found with the FineWebEDU training annotation set in Section~\ref{sec:fwe_training_data}.

\section{German and Chinese Filtering Evaluations}
\label{sec:german_chinese_filter_eval}
Section~\ref{sec:experiments} focuses on French-English bilingual models due to the (i) availability of a sufficient amount of data for training 1.3B models for both curated data (FineWeb2) and common crawl data (Redpajama2), (ii) multiple evaluations both translated from English data and native, and (iii) closeness to English.  In this section, we additionally show that the filtering improves performance for other languages in both monolingual and bilingual models.  

We compare results for French, German, and Chinese languages from FineWeb2 in  Table~\ref{tab:fw2_lang_mono_filter} for monolingual models and Table~\ref{tab:fw2_lang_bi_filter} for bilingual models, and French and German for RedPajamav2\footnote{For German experiments on RedPajamav2, we use a smaller set of data and repeat for two repetitions.  This amount of repetition should not have an effect following \cite{muennighoff2024scaling}. Models get similar performance and improvements as for RedPajamav2 in French.}.  In all cases, we observe an improvement in performance with French and German models being better than training on translated high quality English data.  Note further that for all three models, there is reduced gap in performance between monolingual and bilingual models for all languages. We report that the gap in English performances to a monolingual English model improves by up to $\sim$1\% with filtered data from FineWeb2 in other languages. 

\begin{table}[ht]
\centering
\resizebox{\columnwidth}{!}{
    \begin{tabular}{lccccc}
    \toprule
        Model & RPJ2 FR & RPJ2 DE & FW2 FR & FW2 DE & FW2 ZH  \\
        \midrule
        Base & 48.38 & 48.33 &  51.61 & 49.83 & 51.36\\
        90\% Filter & 50.42 & 49.86 &  54.17  & 52.53 & 53.14\\
        \bottomrule
    \end{tabular}
}
\caption{Comparison of filtering for RPJ2 and FW2 for monolingual models  for Core benchmarks in the native languages.}
\label{tab:fw2_lang_mono_filter}
\end{table}

\begin{table}[ht]
\centering
\resizebox{\columnwidth}{!}{
    \begin{tabular}{lccccc}
    \toprule
        Model & RPJ2 FR & RPJ2 DE &  FW2 FR & FW2 DE & FW2 ZH  \\
        \midrule
        Base & 52.16 & 50.70 &  53.65 & 52.25 & 51.50\\
        90\% Filter & 53.37 & 51.58 &  55.47 & 53.90 & 53.97\\
        \bottomrule
    \end{tabular}
}
\caption{Comparison of filtering for RPJ2 and FW2 for bilingual models for Core benchmarks in the non-english languages.  Models are trained with the respective datasets and FineWebEDU in English.}
\label{tab:fw2_lang_bi_filter}
\end{table}

\section{Multilingual Model Training}
Section~\ref{sec:german_chinese_filter_eval} shows that our filtering approach improves performance for multiple languages, however all models trained are still bilingual English and other language.  We investigate whether improvements hold for multilingual models.  We train a 1.3B parameter model for 400K steps with an equal number of steps per language using the SentenceBert filtered FineWeb2 corpus for Chinese, French, and German, and FineWebEDU as the English corpus.  Results are provided in Table~\ref{tab:1_3B_multilingual} at 200K steps to match the same number of training flops, and 400K steps to match training on the same amount of data per language.  We see that when training on the same amount of data, we attain similar performance to the bilingual models with filtering.  When examining results at the 200K intermediate step, we see slightly lower performance than at 400K steps noting that the multilingual model should be trained for longer, and is trained on less data per language.  These results are consistent with the bilingual results in Sections~\ref{sec:data_quality} and \ref{sec:experiments}.

\begin{table}[ht]
\centering
\resizebox{\columnwidth}{!}{
    \begin{tabular}{lccc}
    \toprule
        Model & Core FR & Core DE & Core ZH  \\
        \midrule
        90\% Filter Bi & 55.47 & 53.90 & 53.97\\
        90\% Filter Multi (200K) & 54.06 & 52.48 & 53.45\\
        90\% Filter Multi (400K) & 54.96 & 53.86 & 54.31\\
        \bottomrule
    \end{tabular}
}
\caption{Comparison of filtering for bilingual and multilingual models for Core benchmarks in the non-english languages.  Models are trained with roughly the same amount of data from each language filtered from FineWeb2 for the same number of steps (100K) each. For Englishd ata, we use FineWebEDU.}
\label{tab:1_3B_multilingual}
\end{table}

\section{Model Scaling}
\label{sec:appendix_model_scaling}

\subsection{Model Scaling Experiments}
\label{sec:model_scaling}
\begin{figure}[ht]
    \centering
    \includegraphics[width=\linewidth]{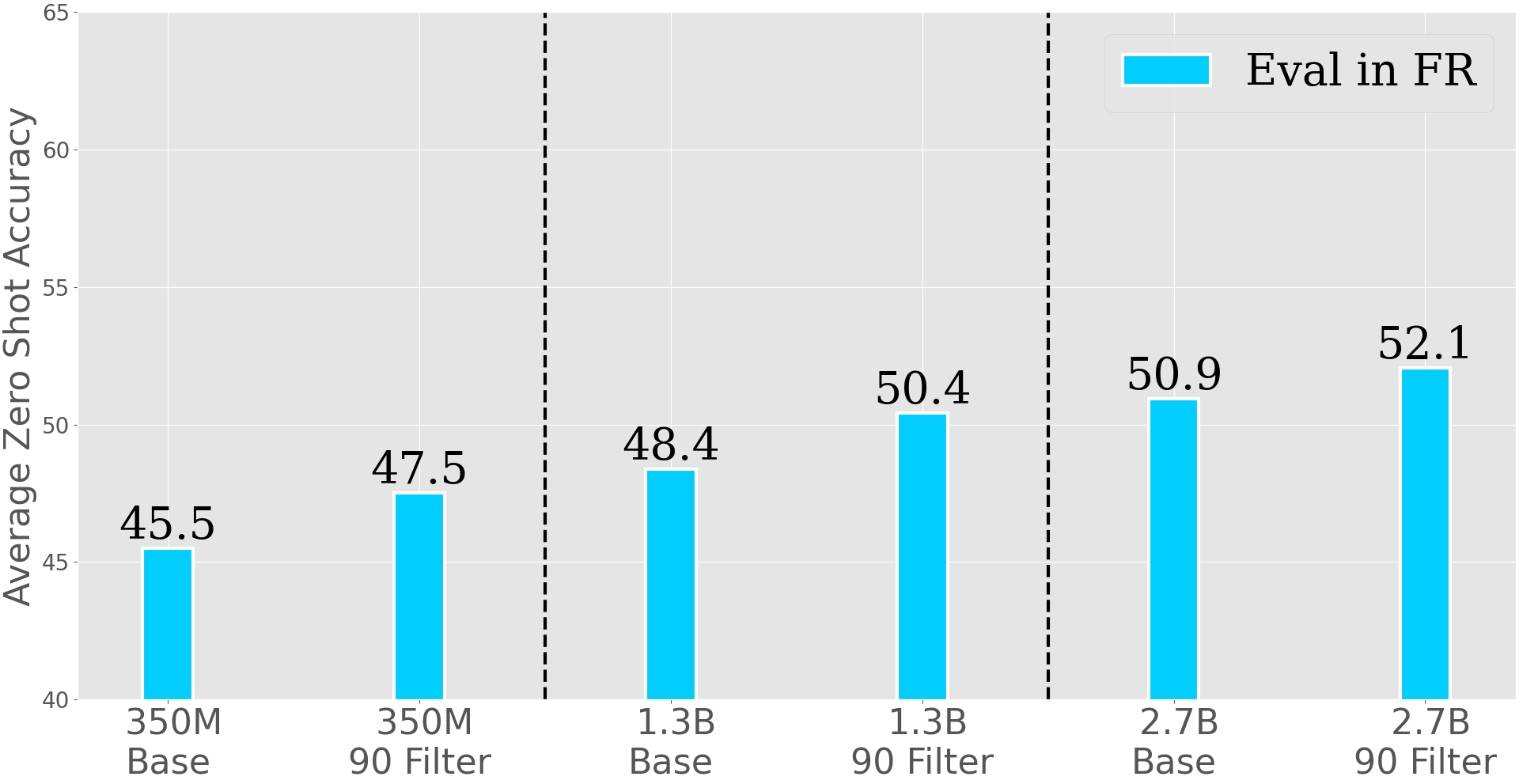}
    \caption{Monolingual model performance comparing filtering on the Core FR benchmarks for various model sizes.}
    \label{fig:mono_model_barplot}
\end{figure}

\begin{figure}[ht]
    \centering
    \includegraphics[width=\linewidth]{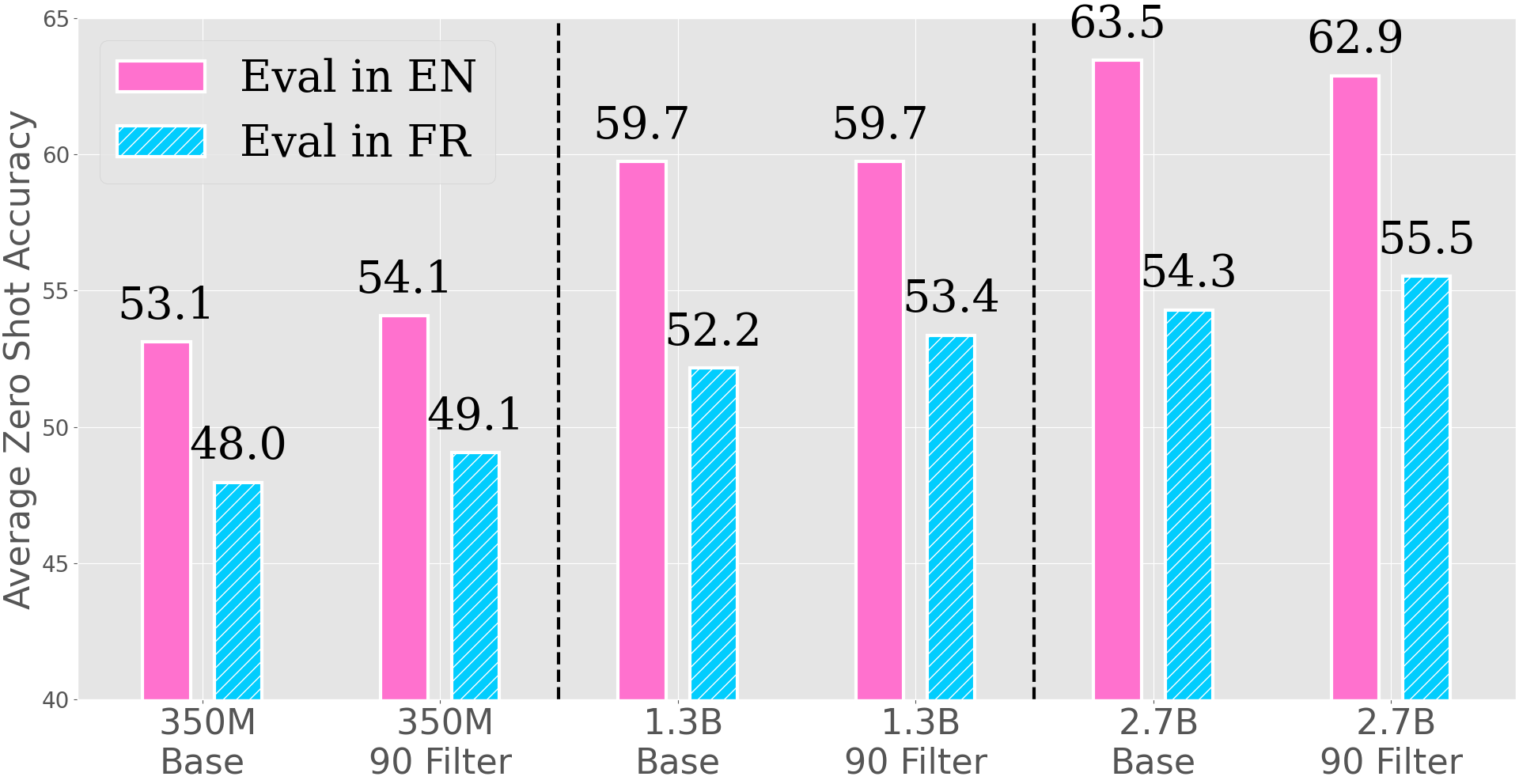}
    \caption{Bilingual model performance comparing filtering on the Core EN and FR benchmarks for various model sizes.}
    \label{fig:bi_model_barplot}
\end{figure}

\paragraph{Methodology:} We investigate to what extent our results are similar across model sizes.  We measure performance at three model sizes: 350M, 1.3B, and 2.7B, and train models for each model size trained on the same data pools for both the base distribution of RedPajamav2 FR and filtered version at 90\% filtering.  Note that the 2.7B model has twice the context length and  sees data for twice as many repetitions.  We report results for the monolingual models in Figure~\ref{fig:mono_model_barplot} and for bilingual performance in Figure~\ref{fig:bi_model_barplot}.   

 \paragraph{Findings:} For monolingual models, we see 2\% improvement for 350M and 1.3B, and 1\% for the 2.7B model.  With more data, it's possible to see greater improvements on the 2.7B model, however we note that we both saturate on the amount of filtered data requiring multiple repetitions, and the scale drops to only 2-3$\times$ Chinchilla vs. $7\times$.  For bilingual performance, we see consistent performance of 1-1.5\% improvement for French evaluations.  However, there is little improvement from filtering on English performance as is similar to prior results.  This is similar to results in Section~\ref{sec:bilingual_data_quality}, and is likely a result of the English FineWebEDU dataset having higher quality data relevant to the downstream evaluations than even filtered RedPajamav2.

\subsection{Comparison for 2.7B Parameter Models}

\begin{figure*}[ht]
    \centering
       \begin{subfigure}{0.3\textwidth}
        \centering
        \includegraphics[width=\textwidth]{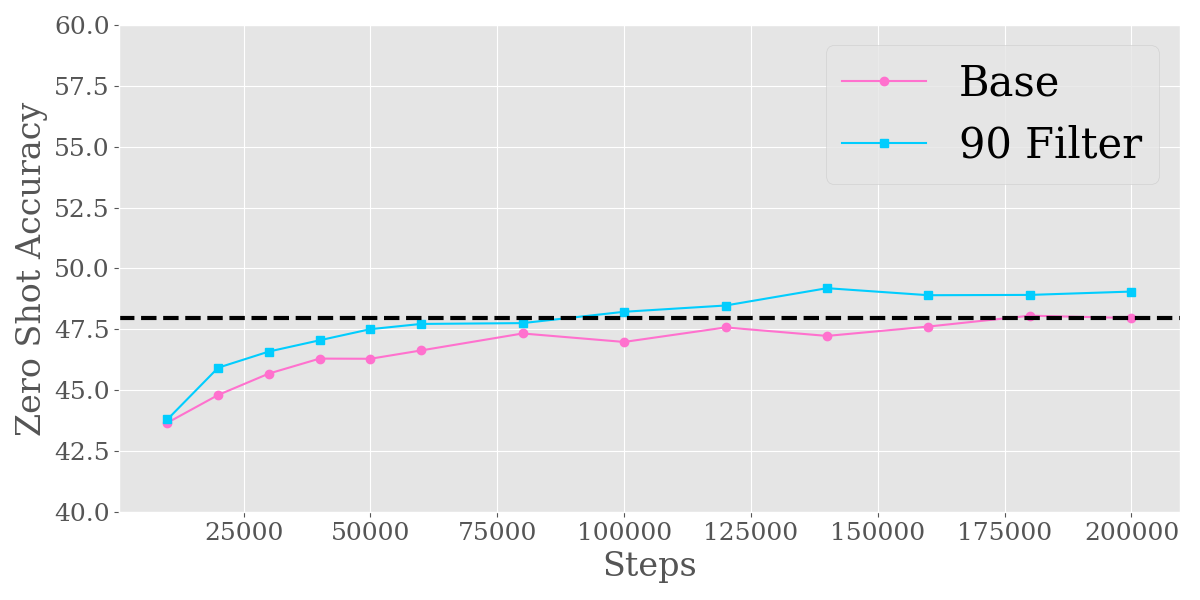}
        \caption{350M}
        \label{fig:350m_scaling}
    \end{subfigure}
    \hfill
    \begin{subfigure}{0.3\textwidth}
        \centering
        \includegraphics[width=\textwidth]{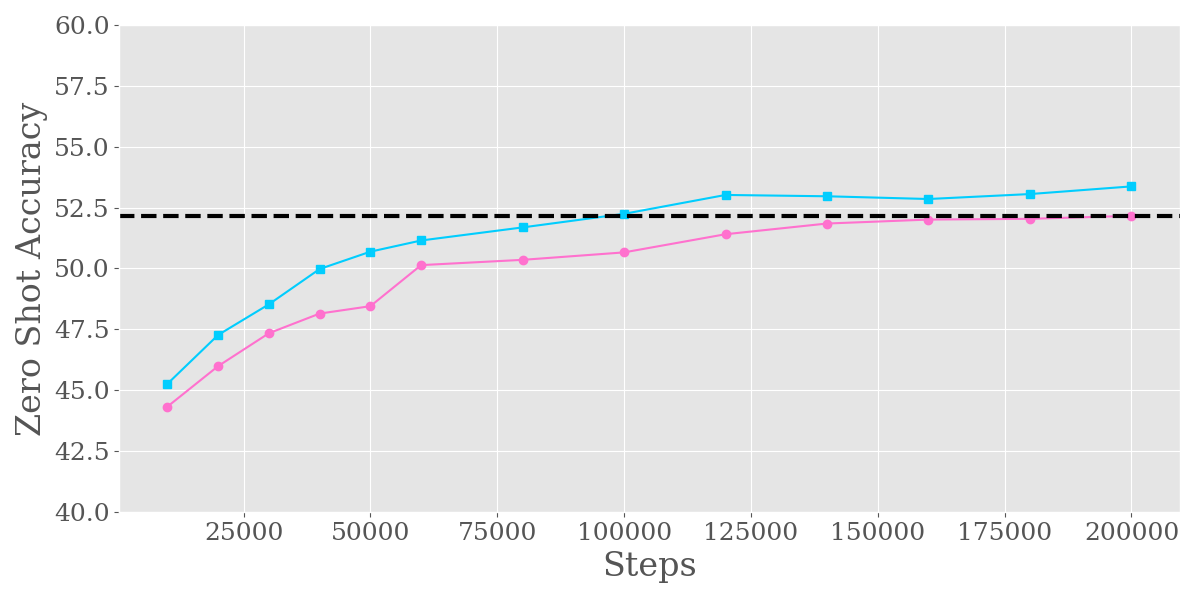}
        \caption{1.3B}
        \label{fig:1_3b_scaling}
    \end{subfigure}
    \hfill
    \begin{subfigure}{0.3\textwidth}
        \centering
        \includegraphics[width=\textwidth]{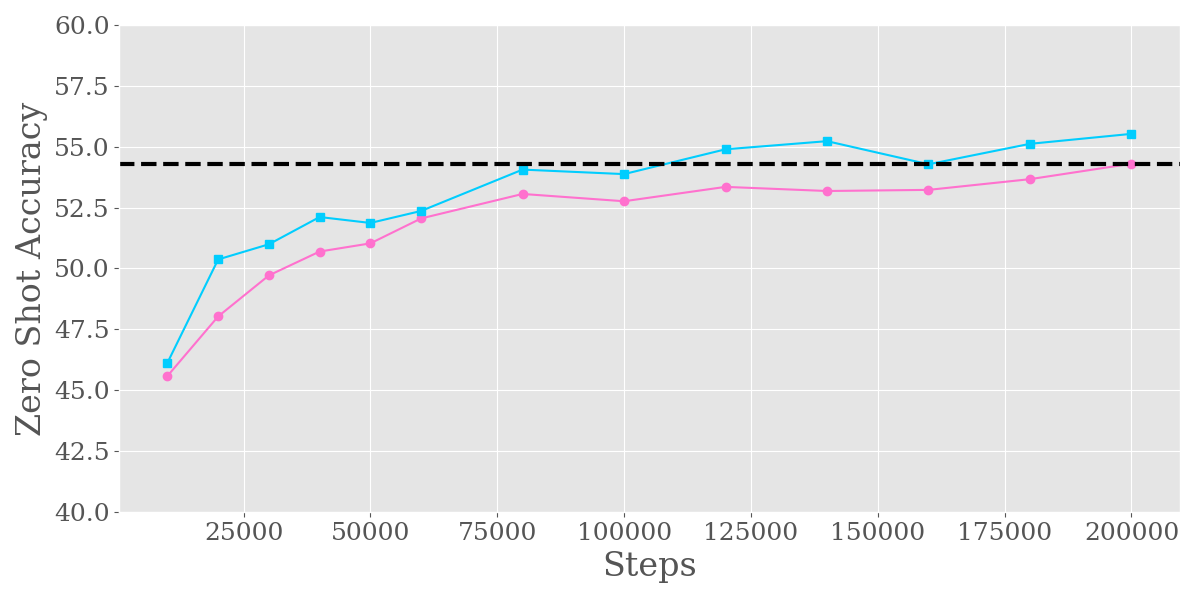}
        \caption{2.7B}
        \label{fig:2_7b_scaling}
    \end{subfigure}
       
    \caption{Performance at intermediate checkpoints during training for different model sizes models for Core FR benchmarks.}
    \label{fig:all_model_scaling}
\end{figure*}

Section~\ref{sec:model_scaling} shows that as we increase model size, performance also increases indicating that filtering improves performance regardless of model size. We first show that this also applies at all intermediate checkpoints where we observe consistent trends regardless of the number of steps in Figure~\ref{fig:all_model_scaling}.  Next, we show comparison with public state-of-the-art multilingual models such as Qwen2.5 3B \cite{yang2024qwen2}, and Helium-1\footnote{\url{https://kyutai.org/2025/04/30/helium.html}} in Table~\ref{tab:3B_fr_comp_public} on the same French evaluation sets from Section~\ref{sec:public_comparison}. At the 2.7B parameter scale, our model trained with filtered RedPajamav2 outperforms the unfiltered model.  We do not train on the FineWeb2 as the amount of filtered data is small and would require over 10x repetition which may impact performance. Performance of our models are lower than Qwen2.5 and Helium-1 models at this scale.  This is because our models are trained with the same data as the smaller models for consistency and comparison leading to more repetitions even for English data, and models such as the Qwen2.5 family are trained on 18T tokens \cite{yang2024qwen2}, which is over 40$\times$ the data used by our models.  We see consistent improvements in early stages of training (first repetition of data), and expect that with more data in other languages for filtering, performance could improve as well.

\begin{table*}
    \centering
\begin{tabular}{l|cc|ccccc>{\columncolor{LightCyan}}c}
    \toprule 
    \textbf{Model} & \textbf{Core EN} & \textbf{MMLU EN} &  \textbf{Core} & \textbf{MMLU} & \textbf{FB-MC} & \textbf{Regional} &  \textbf{NLI}  & \textbf{AVG}\\
    \midrule
    RPJ2 & 63.45 & 35.86  & 54.30 & 30.28 & 61.33 & \underline{34.54} & 40.86 & 44.26\\
    Helium-1 2B & \underline{71.26} & \underline{41.11} & \underline{60.13} & 33.53 &  \textbf{64.28} & 34.51 & \underline{42.15} & \underline{46.92}\\
    Bloom 3B & 58.21 & 31.68 & 52.62 & 28.93 & 60.24 & 31.55 & 40.50 & 42.77\\
    Qwen 3B & \textbf{72.54} & \textbf{44.79} & \textbf{60.95} & 34.96 & \underline{63.52} & \textbf{36.62} & \textbf{52.32} & \textbf{49.67}\\
    RPJ2 90\% & 62.89 & 35.03 & 55.52 & 30.72 & 63.26 & 34.51  & 40.70 & 44.94\\
    \bottomrule
\end{tabular}
\caption{Comparison of different public 2B+ bilingual models in French and English.}
\label{tab:3B_fr_comp_public}
\end{table*}

\label{sec:appendix_3B_public}

\section{Individual Task Accuracies}
\label{sec:indiv_task_accs}

We provide individual accuracies for all models we train with our filtering strategy and evaluate in  Section~\ref{sec:experiments}.  

\subsection{Monolingual French Filter Performance}

This section provides results for individual tasks in the Core benchmark for monolingual French models with varying quality to supplement Figure~\ref{fig:sbert_filter_fr}, and Table~\ref{tab:fw2_rpj_fr}. Results are presented in Table~\ref{tab:indiv_results_mono_fr}.

\begin{table*}
\resizebox{\textwidth}{!}{
    \begin{tabular}{lcccccc>{\columncolor{LightCyan}}c}
\toprule
Model Name & ARC-C & ARC-E & HS & PIQA & SCIQ & WG & AVG \\
\midrule
RPJ2 30\% & $22.18 \pm 1.21$ & $35.06 \pm 0.98$ & $30.81 \pm 0.46$ & $56.37 \pm 1.16$ & $57.30 \pm 1.56$ & $51.54 \pm 1.40$ & $42.21$ \\
RPJ2 60\% & $23.98 \pm 1.25$ & $37.25 \pm 0.99$ & $29.71 \pm 0.46$ & $57.02 \pm 1.16$ & $66.10 \pm 1.50$ & $50.36 \pm 1.41$ & $44.07$ \\
RPJ2 90\% & $24.40 \pm 1.26$ & $37.67 \pm 0.99$ & $31.69 \pm 0.46$ & $56.42 \pm 1.16$ & $61.90 \pm 1.54$ & $50.75 \pm 1.41$ & $43.80$ \\
RPJ2 95\% & $23.46 \pm 1.24$ & $37.92 \pm 1.00$ & $30.88 \pm 0.46$ & $57.89 \pm 1.15$ & $59.10 \pm 1.56$ & $50.51 \pm 1.41$ & $43.29$ \\
FW2 30\% & $22.95 \pm 1.23$ & $37.42 \pm 0.99$ & $33.89 \pm 0.47$ & $62.30 \pm 1.13$ & $62.40 \pm 1.53$ & $51.70 \pm 1.40$ & $45.11$ \\
FW2 60\% & $24.66 \pm 1.26$ & $38.05 \pm 1.00$ & $34.90 \pm 0.48$ & $61.97 \pm 1.13$ & $60.70 \pm 1.55$ & $50.04 \pm 1.41$ & $45.05$ \\
FW2 90\% & $26.79 \pm 1.29$ & $43.43 \pm 1.02$ & $38.89 \pm 0.49$ & $63.33 \pm 1.12$ & $65.80 \pm 1.50$ & $52.57 \pm 1.40$ & $48.47$ \\
\midrule
RPJ2 30\% & $25.72 \pm 1.29$ & $37.34 \pm 1.02$ & $42.16 \pm 0.51$ & $63.33 \pm 1.12$ & $60.86 \pm 1.58$ & $51.93 \pm 1.43$ & $46.89$ \\
RPJ2 60\% & $27.03 \pm 1.31$ & $42.36 \pm 1.04$ & $43.02 \pm 0.51$ & $63.38 \pm 1.12$ & $64.74 \pm 1.55$ & $52.43 \pm 1.43$ & $48.83$ \\
RPJ2 90\% & $28.51 \pm 1.33$ & $45.84 \pm 1.05$ & $45.05 \pm 0.51$ & $65.13 \pm 1.11$ & $65.06 \pm 1.55$ & $52.92 \pm 1.43$ & $50.42$ \\
RPJ2 95\% & $27.29 \pm 1.32$ & $44.52 \pm 1.04$ & $44.27 \pm 0.51$ & $64.85 \pm 1.11$ & $61.18 \pm 1.58$ & $53.50 \pm 1.43$ & $49.27$ \\
FW2 30\% & $25.02 \pm 1.28$ & $39.41 \pm 1.03$ & $45.20 \pm 0.52$ & $66.05 \pm 1.10$ & $60.65 \pm 1.58$ & $54.07 \pm 1.43$ & $48.40$ \\
FW2 60\% & $27.55 \pm 1.32$ & $39.63 \pm 1.03$ & $47.41 \pm 0.52$ & $66.32 \pm 1.10$ & $61.18 \pm 1.58$ & $53.00 \pm 1.43$ & $49.18$ \\
FW2 90\% & $30.95 \pm 1.37$ & $48.88 \pm 1.05$ & $52.55 \pm 0.52$ & $68.77 \pm 1.08$ & $68.31 \pm 1.51$ & $55.56 \pm 1.43$ & $54.17$ \\
\bottomrule
\end{tabular}
}

\caption{Evaluation of 1.3B parameter monolingual French models on general understanding tasks for English (top) and French (bottom) with varying quality. All evaluations are zero-shot.}
\label{tab:indiv_results_mono_fr}
\end{table*}

\subsection{Bilingual French Filter Performance }

This section provides results for individual tasks in the Core benchmark for monolingual French models with varying quality to supplement Figure~\ref{fig:filter_bilingual}. Results are presented in Table~\ref{tab:indiv_results_bi_fr}.

\begin{table*}
\resizebox{\textwidth}{!}{
    \begin{tabular}{lcccccc>{\columncolor{LightCyan}}c}
\toprule
Model Name & ARC-C & ARC-E & HS & PIQA & SCIQ & WG & AVG \\
\midrule
FWE FR & $32.00 \pm 1.36$ & $53.45 \pm 1.02$ & $42.53 \pm 0.49$ & $65.23 \pm 1.11$ & $72.10 \pm 1.42$ & $54.78 \pm 1.40$ & $53.35$ \\
FWE EN & $37.71 \pm 1.42$ & $66.08 \pm 0.97$ & $56.84 \pm 0.49$ & $73.78 \pm 1.03$ & $81.20 \pm 1.24$ & $57.38 \pm 1.39$ & $62.16$ \\
FWE EN RPJ2 30\% & $32.76 \pm 1.37$ & $62.08 \pm 1.00$ & $51.66 \pm 0.50$ & $71.33 \pm 1.06$ & $78.80 \pm 1.29$ & $57.14 \pm 1.39$ & $58.96$ \\
FWE EN RPJ2 60\% & $33.28 \pm 1.38$ & $62.29 \pm 0.99$ & $52.09 \pm 0.50$ & $72.09 \pm 1.05$ & $81.50 \pm 1.23$ & $56.43 \pm 1.39$ & $59.61$ \\
FWE EN RPJ2 90\% & $34.90 \pm 1.39$ & $62.46 \pm 0.99$ & $52.75 \pm 0.50$ & $71.33 \pm 1.06$ & $80.70 \pm 1.25$ & $56.27 \pm 1.39$ & $59.73$ \\
FWE EN FW2 30\% & $34.13 \pm 1.39$ & $60.40 \pm 1.00$ & $53.07 \pm 0.50$ & $72.42 \pm 1.04$ & $79.30 \pm 1.28$ & $56.99 \pm 1.39$ & $59.38$ \\
FWE EN FW2 60\% & $33.53 \pm 1.38$ & $61.53 \pm 1.00$ & $54.48 \pm 0.50$ & $72.31 \pm 1.04$ & $80.10 \pm 1.26$ & $56.12 \pm 1.39$ & $59.68$ \\
FWE EN FW2 90\% & $36.26 \pm 1.40$ & $62.37 \pm 0.99$ & $55.74 \pm 0.50$ & $73.50 \pm 1.03$ & $81.30 \pm 1.23$ & $57.22 \pm 1.39$ & $61.07$ \\
\midrule 
FWE FR & $34.26 \pm 1.40$ & $53.28 \pm 1.05$ & $48.13 \pm 0.52$ & $63.17 \pm 1.13$ & $70.41 \pm 1.48$ & $56.46 \pm 1.42$ & $54.28$ \\
FWE EN & $25.11 \pm 1.28$ & $32.94 \pm 0.99$ & $30.64 \pm 0.48$ & $52.01 \pm 1.17$ & $62.22 \pm 1.57$ & $50.95 \pm 1.43$ & $42.31$ \\
FWE EN RPJ2 30\% & $29.56 \pm 1.35$ & $45.88 \pm 1.05$ & $46.13 \pm 0.52$ & $63.66 \pm 1.12$ & $66.84 \pm 1.53$ & $53.58 \pm 1.43$ & $50.94$ \\
FWE EN RPJ2 60\% & $30.95 \pm 1.37$ & $47.95 \pm 1.05$ & $47.44 \pm 0.52$ & $66.00 \pm 1.11$ & $67.58 \pm 1.52$ & $54.73 \pm 1.43$ & $52.44$ \\
FWE EN RPJ2 90\% & $31.12 \pm 1.37$ & $49.45 \pm 1.05$ & $48.89 \pm 0.52$ & $65.94 \pm 1.11$ & $69.57 \pm 1.49$ & $55.23 \pm 1.43$ & $53.37$ \\
FWE EN FW2 30\% & $29.99 \pm 1.35$ & $45.40 \pm 1.04$ & $47.86 \pm 0.52$ & $66.54 \pm 1.10$ & $67.47 \pm 1.52$ & $53.91 \pm 1.43$ & $51.86$ \\
FWE EN FW2 60\% & $28.86 \pm 1.34$ & $45.31 \pm 1.04$ & $50.07 \pm 0.52$ & $66.70 \pm 1.10$ & $67.47 \pm 1.52$ & $55.80 \pm 1.43$ & $52.37$ \\
FWE EN FW2 90\% & $32.61 \pm 1.38$ & $50.15 \pm 1.05$ & $53.21 \pm 0.52$ & $69.26 \pm 1.08$ & $71.14 \pm 1.47$ & $56.46 \pm 1.42$ & $55.47$ \\
\bottomrule
\end{tabular}
}

\caption{Evaluation of 1.3B parameter models on general understanding tasks for English (top) and French (bottom) with varying quality. All evaluations are zero-shot.}
\label{tab:indiv_results_bi_fr}
\end{table*}

\subsection{Filter Performance Comparison with FineWeb 2 HQ}

We report performance for individual tasks comparing FineWeb2 with our filtering and FineWeb2 HQ \cite{messmer2025enhancing}. Table~\ref{tab:indiv_fw2_hq_mono} shows results for monolingual models and Table~\ref{tab:indiv_fw2_hq_bi} for bilingual models with FineWebEDU in English.

\begin{table*}
\resizebox{\textwidth}{!}{
    \begin{tabular}{lcccccc>{\columncolor{LightCyan}}c}
\toprule
Model Name & ARC-C & ARC-E & HS & PIQA & SCIQ & WG & AVG \\
\midrule
FW2 FR & $25.77 \pm 1.28$ & $39.44 \pm 1.00$ & $37.21 \pm 0.48$ & $64.09 \pm 1.12$ & $65.90 \pm 1.50$ & $52.41 \pm 1.40$ & $47.47$ \\
FW2 FR 90\% & $26.79 \pm 1.29$ & $43.43 \pm 1.02$ & $38.89 \pm 0.49$ & $63.33 \pm 1.12$ & $65.80 \pm 1.50$ & $52.57 \pm 1.40$ & $48.47$ \\
FW2 FR HQ & $27.82 \pm 1.31$ & $44.19 \pm 1.02$ & $37.40 \pm 0.48$ & $63.55 \pm 1.12$ & $69.10 \pm 1.46$ & $52.25 \pm 1.40$ & $49.05$ \\
\midrule 
FW2 FR & $28.95 \pm 1.34$ & $43.28 \pm 1.04$ & $48.58 \pm 0.52$ & $67.85 \pm 1.09$ & $66.84 \pm 1.53$ & $54.16 \pm 1.43$ & $51.61$ \\
FW2 FR 90%
FW2 FR HQ & $30.69 \pm 1.36$ & $46.46 \pm 1.05$ & $49.87 \pm 0.52$ & $68.17 \pm 1.09$ & $67.58 \pm 1.52$ & $54.90 \pm 1.43$ & $52.94$ \\
\bottomrule
\end{tabular}
}
\caption{Evaluation of 1.3B parameter models on general understanding tasks for English (top) and French (bottom) with varying quality. Models are trained on FineWeb2 (FW2) or FineWeb 2 HQ (FW2 HQ) in French with and without filtering. All evaluations are zero-shot.}
\label{tab:indiv_fw2_hq_mono}
\end{table*}

\begin{table*}
\resizebox{\textwidth}{!}{
    \begin{tabular}{lcccccc>{\columncolor{LightCyan}}c}
\toprule
Model Name & ARC-C & ARC-E & HS & PIQA & SCIQ & WG & AVG \\
\midrule
FWE EN FW2 FR & $34.64 \pm 1.39$ & $62.25 \pm 0.99$ & $54.31 \pm 0.50$ & $72.31 \pm 1.04$ & $80.30 \pm 1.26$ & $57.77 \pm 1.39$ & $60.26$ \\
FWE EN + FW2 FR 90\% & $36.26 \pm 1.40$ & $62.37 \pm 0.99$ & $55.74 \pm 0.50$ & $73.50 \pm 1.03$ & $81.30 \pm 1.23$ & $57.22 \pm 1.39$ & $61.07$ \\
FWE EN FW2 FR HQ & $35.92 \pm 1.40$ & $64.35 \pm 0.98$ & $54.57 \pm 0.50$ & $73.34 \pm 1.03$ & $81.80 \pm 1.22$ & $57.22 \pm 1.39$ & $61.20$ \\
\midrule 
FWE EN FW2 FR & $31.21 \pm 1.37$ & $48.75 \pm 1.05$ & $50.27 \pm 0.52$ & $68.12 \pm 1.09$ & $70.30 \pm 1.48$ & $53.25 \pm 1.43$ & $53.65$ \\
FWE EN + FW2 FR 90\% & $32.61 \pm 1.38$ & $50.15 \pm 1.05$ & $53.21 \pm 0.52$ & $69.26 \pm 1.08$ & $71.14 \pm 1.47$ & $56.46 \pm 1.42$ & $55.47$ \\
FWE EN FW2 FR HQ & $32.52 \pm 1.38$ & $51.74 \pm 1.05$ & $51.31 \pm 0.52$ & $68.55 \pm 1.08$ & $71.46 \pm 1.46$ & $53.50 \pm 1.43$ & $54.85$ \\
\bottomrule
\end{tabular}
}
\caption{Evaluation of 1.3B parameter models on general understanding tasks for English (top) and French (bottom) with varying quality. Models are trained on FineWeb2 (FW2) or FineWeb 2 HQ (FW2 HQ) in French with and without filtering and FineWebEDU (FWE) in English. All evaluations are zero-shot.}
\label{tab:indiv_fw2_hq_bi}
\end{table*}

\subsection{Filter Performance Across Languages}

This section expands on results for Tables~\ref{tab:fw2_lang_mono_filter}-\ref{tab:fw2_lang_bi_filter}.  For monolingual models, Table~\ref{tab:indiv_fr_mono} presents individual task accuracy for French, Table~\ref{tab:indiv_de_mono} for German, and Table~\ref{tab:indiv_zh_mono} for Chinese.

individual task accuracies for bilingual models with FineWebEDU in English are included in  Table~\ref{tab:indiv_fr_bi}  for French, Table~\ref{tab:indiv_de_bi} for German, and Table~\ref{tab:indiv_zh_bi} for Chinese.

\begin{table*}
\resizebox{\textwidth}{!}{
    \begin{tabular}{lcccccc>{\columncolor{LightCyan}}c}
\toprule
Model Name & ARC-C & ARC-E & HS & PIQA & SCIQ & WG & AVG \\
\midrule
RPJ2 FR & $23.46 \pm 1.24$ & $36.78 \pm 0.99$ & $32.03 \pm 0.47$ & $56.86 \pm 1.16$ & $64.80 \pm 1.51$ & $49.96 \pm 1.41$ & $43.98$ \\
RPJ2 FR 90\% & $24.40 \pm 1.26$ & $37.67 \pm 0.99$ & $31.69 \pm 0.46$ & $56.42 \pm 1.16$ & $61.90 \pm 1.54$ & $50.75 \pm 1.41$ & $43.80$ \\
FW2 FR & $25.77 \pm 1.28$ & $39.44 \pm 1.00$ & $37.21 \pm 0.48$ & $64.09 \pm 1.12$ & $65.90 \pm 1.50$ & $52.41 \pm 1.40$ & $47.47$ \\
FW2 FR 90\% & $26.79 \pm 1.29$ & $43.43 \pm 1.02$ & $38.89 \pm 0.49$ & $63.33 \pm 1.12$ & $65.80 \pm 1.50$ & $52.57 \pm 1.40$ & $48.47$ \\
\midrule
RPJ2 FR & $24.85 \pm 1.28$ & $41.35 \pm 1.03$ & $43.52 \pm 0.51$ & $64.04 \pm 1.12$ & $62.54 \pm 1.57$ & $53.99 \pm 1.43$ & $48.38$ \\
RPJ2 FR 90\% & $28.51 \pm 1.33$ & $45.84 \pm 1.05$ & $45.05 \pm 0.51$ & $65.13 \pm 1.11$ & $65.06 \pm 1.55$ & $52.92 \pm 1.43$ & $50.42$ \\
FW2 FR & $28.95 \pm 1.34$ & $43.28 \pm 1.04$ & $48.58 \pm 0.52$ & $67.85 \pm 1.09$ & $66.84 \pm 1.53$ & $54.16 \pm 1.43$ & $51.61$ \\
FW2 FR 90\% & $30.95 \pm 1.37$ & $48.88 \pm 1.05$ & $52.55 \pm 0.52$ & $68.77 \pm 1.08$ & $68.31 \pm 1.51$ & $55.56 \pm 1.43$ & $54.17$ \\
\bottomrule
\end{tabular}
}
\caption{Evaluation of 1.3B parameter models on general understanding tasks for English (top) and French (bottom) with varying quality. Models are trained on RedPajamav2 (RPJ2) or FineWeb2 (FW2) in French with and without filtering. All evaluations are zero-shot.}
\label{tab:indiv_fr_mono}
\end{table*}

\begin{table*}
\resizebox{\textwidth}{!}{
    \begin{tabular}{lcccccc>{\columncolor{LightCyan}}c}
\toprule
Model Name & ARC-C & ARC-E & HS & PIQA & SCIQ & WG & AVG \\
\midrule
RPJ2 DE & $24.23 \pm 1.25$ & $37.92 \pm 1.00$ & $31.69 \pm 0.46$ & $58.16 \pm 1.15$ & $63.30 \pm 1.52$ & $49.64 \pm 1.41$ & $44.16$ \\
RPJ2 DE 90\% & $24.49 \pm 1.26$ & $38.59 \pm 1.00$ & $31.53 \pm 0.46$ & $58.22 \pm 1.15$ & $64.70 \pm 1.51$ & $51.38 \pm 1.40$ & $44.82$ \\
FW2 DE & $24.91 \pm 1.26$ & $35.10 \pm 0.98$ & $36.58 \pm 0.48$ & $61.70 \pm 1.13$ & $67.10 \pm 1.49$ & $52.01 \pm 1.40$ & $46.23$ \\
FW2 DE 90\% & $24.57 \pm 1.26$ & $42.59 \pm 1.01$ & $37.86 \pm 0.48$ & $63.44 \pm 1.12$ & $66.70 \pm 1.49$ & $50.28 \pm 1.41$ & $47.57$ \\
\midrule 
RPJ2 DE & $27.88 \pm 1.33$ & $42.74 \pm 1.04$ & $40.02 \pm 0.51$ & $61.64 \pm 1.13$ & $67.89 \pm 1.52$ & $52.28 \pm 1.45$ & $48.74$ \\
RPJ2 DE 90\% & $29.99 \pm 1.36$ & $45.22 \pm 1.05$ & $41.29 \pm 0.51$ & $63.06 \pm 1.13$ & $71.16 \pm 1.47$ & $51.27 \pm 1.45$ & $50.33$ \\
FW2 DE & $27.79 \pm 1.33$ & $39.07 \pm 1.03$ & $43.58 \pm 0.51$ & $64.20 \pm 1.12$ & $69.79 \pm 1.49$ & $54.56 \pm 1.45$ & $49.83$ \\
FW2 DE 90\% & $29.11 \pm 1.35$ & $47.48 \pm 1.05$ & $47.78 \pm 0.52$ & $66.16 \pm 1.10$ & $69.26 \pm 1.50$ & $55.41 \pm 1.45$ & $52.53$ \\
\bottomrule
\end{tabular}
}
\caption{Evaluation of 1.3B parameter models on general understanding tasks for English (top) and German (bottom) with varying quality. Models are trained on RedPajamav2 (RPJ2) or FineWeb2 (FW2) in German with and without filtering. All evaluations are zero-shot.}
\label{tab:indiv_de_mono}
\end{table*}

\begin{table*}
\resizebox{\textwidth}{!}{
    \begin{tabular}{lcccccc>{\columncolor{LightCyan}}c}
\toprule
Model Name & ARC-C & ARC-E & HS & PIQA & SCIQ & WG & AVG \\
\midrule
FW2 ZH & $22.18 \pm 1.21$ & $36.62 \pm 0.99$ & $30.09 \pm 0.46$ & $57.07 \pm 1.15$ & $64.20 \pm 1.52$ & $52.41 \pm 1.40$ & $43.76$ \\
FW2 ZH 90\% & $21.76 \pm 1.21$ & $36.36 \pm 0.99$ & $30.73 \pm 0.46$ & $58.16 \pm 1.15$ & $64.30 \pm 1.52$ & $53.04 \pm 1.40$ & $44.06$ \\
\midrule 
FW2 ZH & $28.62 \pm 1.34$ & $47.56 \pm 1.05$ & $40.25 \pm 0.51$ & $61.92 \pm 1.13$ & $76.00 \pm 1.35$ & $53.82 \pm 1.53$ & $51.36$ \\
FW2 ZH 90\% & $30.02 \pm 1.35$ & $52.27 \pm 1.05$ & $43.05 \pm 0.51$ & $63.33 \pm 1.12$ & $77.60 \pm 1.32$ & $52.60 \pm 1.54$ & $53.14$ \\
\bottomrule
\end{tabular}
}
\caption{Evaluation of 1.3B parameter models on general understanding tasks for English (top) and Chinese (bottom) with varying quality. Models are trained on RedPajamav2 (RPJ2) or FineWeb2 (FW2) in Chinese with and without filtering. All evaluations are zero-shot.}
\label{tab:indiv_zh_mono}
\end{table*}

\begin{table*}
\resizebox{\textwidth}{!}{
    \begin{tabular}{lcccccc>{\columncolor{LightCyan}}c}
\toprule
Model Name & ARC-C & ARC-E & HS & PIQA & SCIQ & WG & AVG \\
\midrule
FWE EN RPJ2 FR & $35.32 \pm 1.40$ & $61.57 \pm 1.00$ & $52.56 \pm 0.50$ & $70.73 \pm 1.06$ & $81.60 \pm 1.23$ & $56.67 \pm 1.39$ & $59.74$ \\
FWE EN RPJ2 FR 90\% & $34.90 \pm 1.39$ & $62.46 \pm 0.99$ & $52.75 \pm 0.50$ & $71.33 \pm 1.06$ & $80.70 \pm 1.25$ & $56.27 \pm 1.39$ & $59.73$ \\
FWE EN FW2 FR & $34.64 \pm 1.39$ & $62.25 \pm 0.99$ & $54.31 \pm 0.50$ & $72.31 \pm 1.04$ & $80.30 \pm 1.26$ & $57.77 \pm 1.39$ & $60.26$ \\
FWE EN + FW2 FR 90\% & $36.26 \pm 1.40$ & $62.37 \pm 0.99$ & $55.74 \pm 0.50$ & $73.50 \pm 1.03$ & $81.30 \pm 1.23$ & $57.22 \pm 1.39$ & $61.07$ \\
\midrule
FWE EN RPJ2 FR & $30.34 \pm 1.36$ & $47.16 \pm 1.05$ & $47.37 \pm 0.52$ & $64.58 \pm 1.12$ & $69.67 \pm 1.49$ & $53.83 \pm 1.43$ & $52.16$ \\
FWE EN RPJ2 FR 90\% & $31.12 \pm 1.37$ & $49.45 \pm 1.05$ & $48.89 \pm 0.52$ & $65.94 \pm 1.11$ & $69.57 \pm 1.49$ & $55.23 \pm 1.43$ & $53.37$ \\
FWE EN FW2 FR & $31.21 \pm 1.37$ & $48.75 \pm 1.05$ & $50.27 \pm 0.52$ & $68.12 \pm 1.09$ & $70.30 \pm 1.48$ & $53.25 \pm 1.43$ & $53.65$ \\
FWE EN + FW2 FR 90\% & $32.61 \pm 1.38$ & $50.15 \pm 1.05$ & $53.21 \pm 0.52$ & $69.26 \pm 1.08$ & $71.14 \pm 1.47$ & $56.46 \pm 1.42$ & $55.47$ \\
\bottomrule
\end{tabular}
}
\caption{Evaluation of 1.3B parameter models on general understanding tasks for English (top) and French (bottom) with varying quality. Models are trained on FineWebEDU (FWE) in English and RedPajamav2 (RPJ2) or FineWeb2 (FW2) in French with and without filtering. All evaluations are zero-shot.}
\label{tab:indiv_fr_bi}
\end{table*}

\begin{table*}
\resizebox{\textwidth}{!}{
    \begin{tabular}{lcccccc>{\columncolor{LightCyan}}c}
\toprule
Model Name & ARC-C & ARC-E & HS & PIQA & SCIQ & WG & AVG \\
\midrule
FWE EN RPJ2 DE & $35.32 \pm 1.40$ & $61.53 \pm 1.00$ & $51.43 \pm 0.50$ & $70.62 \pm 1.06$ & $82.00 \pm 1.22$ & $54.85 \pm 1.40$ & $59.29$ \\
FWE EN RPJ2 DE 90\% & $33.70 \pm 1.38$ & $61.95 \pm 1.00$ & $51.36 \pm 0.50$ & $70.78 \pm 1.06$ & $82.30 \pm 1.21$ & $55.49 \pm 1.40$ & $59.26$ \\
FWE EN FW2 DE & $33.28 \pm 1.38$ & $59.60 \pm 1.01$ & $53.17 \pm 0.50$ & $72.20 \pm 1.05$ & $80.70 \pm 1.25$ & $56.75 \pm 1.39$ & $59.28$ \\
FWE EN FW2 DE 90\% & $35.84 \pm 1.40$ & $63.89 \pm 0.99$ & $54.93 \pm 0.50$ & $71.55 \pm 1.05$ & $82.80 \pm 1.19$ & $56.12 \pm 1.39$ & $60.85$ \\
\midrule 
FWE EN RPJ2 DE & $29.29 \pm 1.35$ & $46.90 \pm 1.05$ & $42.49 \pm 0.51$ & $62.13 \pm 1.13$ & $71.16 \pm 1.47$ & $53.38 \pm 1.45$ & $50.89$ \\
FWE EN RPJ2 DE 90\% & $30.17 \pm 1.36$ & $49.69 \pm 1.05$ & $43.95 \pm 0.51$ & $62.89 \pm 1.13$ & $72.00 \pm 1.46$ & $53.21 \pm 1.45$ & $51.98$ \\
FWE EN FW2 DE & $29.46 \pm 1.35$ & $47.35 \pm 1.05$ & $45.20 \pm 0.51$ & $65.45 \pm 1.11$ & $72.00 \pm 1.46$ & $54.05 \pm 1.45$ & $52.25$ \\
FWE EN FW2 DE 90\% & $30.61 \pm 1.37$ & $52.30 \pm 1.05$ & $48.54 \pm 0.52$ & $65.18 \pm 1.11$ & $73.47 \pm 1.43$ & $53.29 \pm 1.45$ & $53.90$ \\
\bottomrule
\end{tabular}
}
\caption{Evaluation of 1.3B parameter models on general understanding tasks for English (top) and German (bottom) with varying quality. Models are trained on FineWebEDU (FWE) in English and RedPajamav2 (RPJ2) or FineWeb2 (FW2) in German with and without filtering. All evaluations are zero-shot.}
\label{tab:indiv_de_bi}
\end{table*}

\begin{table*}
\resizebox{\textwidth}{!}{
    \begin{tabular}{lcccccc>{\columncolor{LightCyan}}c}
\toprule
Model Name & ARC-C & ARC-E & HS & PIQA & SCIQ & WG & AVG \\
\midrule
FWE EN FW2 ZH & $33.96 \pm 1.38$ & $60.98 \pm 1.00$ & $49.10 \pm 0.50$ & $70.78 \pm 1.06$ & $77.80 \pm 1.31$ & $55.72 \pm 1.40$ & $58.06$ \\
FWE EN FW2 ZH 90\% & $32.76 \pm 1.37$ & $62.04 \pm 1.00$ & $50.32 \pm 0.50$ & $70.62 \pm 1.06$ & $80.70 \pm 1.25$ & $55.56 \pm 1.40$ & $58.67$ \\
\midrule 
FWE EN FW2 ZH & $29.23 \pm 1.34$ & $49.14 \pm 1.05$ & $40.15 \pm 0.51$ & $61.75 \pm 1.13$ & $78.30 \pm 1.30$ & $50.42 \pm 1.54$ & $51.50$ \\
FWE EN FW2 ZH 90\% & $29.93 \pm 1.35$ & $54.34 \pm 1.05$ & $43.46 \pm 0.51$ & $63.55 \pm 1.12$ & $79.40 \pm 1.28$ & $53.16 \pm 1.53$ & $53.97$ \\
\bottomrule
\end{tabular}
}
\caption{Evaluation of 1.3B parameter models on general understanding tasks for English (top) and Chinese (bottom) with varying quality. Models are trained on FineWebEDU (FWE) in English and FineWeb2 (FW2) in Chinese with and without filtering. All evaluations are zero-shot.}
\label{tab:indiv_zh_bi}
\end{table*}

\subsection{Filter Performance for Varying Model Sizes}

We expand our results for Figure~\ref{fig:bi_model_barplot} showing accuracy for individual tasks in the Core tasks for different model sizes for bilingual English-French models.  All models are trained with FineWebEDU (FWE) as the English data and RedPajama2 (RPJ2) as the French data with and without filtering. Results are shown in Table~\ref{tab:indiv_model_scaling_bi}.

\begin{table*}
\resizebox{\textwidth}{!}{
    \begin{tabular}{lcccccc>{\columncolor{LightCyan}}c}
\toprule
Model Name & ARC-C & ARC-E & HS & PIQA & SCIQ & WG & AVG \\
\midrule
350M FWE EN RPJ2 FR & $27.47 \pm 1.30$ & $53.83 \pm 1.02$ & $43.02 \pm 0.49$ & $67.08 \pm 1.10$ & $75.00 \pm 1.37$ & $52.33 \pm 1.40$ & $53.12$ \\
350M FWE EN RPJ2 FR 90\% & $29.18 \pm 1.33$ & $56.90 \pm 1.02$ & $42.92 \pm 0.49$ & $68.12 \pm 1.09$ & $75.40 \pm 1.36$ & $52.01 \pm 1.40$ & $54.09$ \\
1.3B FWE EN RPJ2 FR & $35.32 \pm 1.40$ & $61.57 \pm 1.00$ & $52.56 \pm 0.50$ & $70.73 \pm 1.06$ & $81.60 \pm 1.23$ & $56.67 \pm 1.39$ & $59.74$ \\
1.3B FWE EN RPJ2 FR 90\% & $34.90 \pm 1.39$ & $62.46 \pm 0.99$ & $52.75 \pm 0.50$ & $71.33 \pm 1.06$ & $80.70 \pm 1.25$ & $56.27 \pm 1.39$ & $59.73$ \\
2.7B FWE EN RPJ2 FR & $38.65 \pm 1.42$ & $68.60 \pm 0.95$ & $57.82 \pm 0.49$ & $73.12 \pm 1.03$ & $84.20 \pm 1.15$ & $58.33 \pm 1.39$ & $63.45$ \\
2.7B FWE EN RPJ2 FR 90\% & $38.23 \pm 1.42$ & $66.08 \pm 0.97$ & $57.04 \pm 0.49$ & $73.18 \pm 1.03$ & $83.60 \pm 1.17$ & $59.19 \pm 1.38$ & $62.89$ \\
\midrule 
350M FWE EN RPJ2 FR & $26.85 \pm 1.31$ & $42.23 \pm 1.04$ & $39.69 \pm 0.51$ & $62.02 \pm 1.13$ & $64.43 \pm 1.55$ & $52.59 \pm 1.43$ & $47.97$ \\
350M FWE EN RPJ2 FR 90\% & $28.25 \pm 1.33$ & $45.22 \pm 1.04$ & $40.93 \pm 0.51$ & $62.95 \pm 1.13$ & $64.22 \pm 1.55$ & $52.76 \pm 1.43$ & $49.05$ \\
1.3B FWE EN RPJ2 FR & $30.34 \pm 1.36$ & $47.16 \pm 1.05$ & $47.37 \pm 0.52$ & $64.58 \pm 1.12$ & $69.67 \pm 1.49$ & $53.83 \pm 1.43$ & $52.16$ \\
1.3B FWE EN RPJ2 FR 90\% & $31.12 \pm 1.37$ & $49.45 \pm 1.05$ & $48.89 \pm 0.52$ & $65.94 \pm 1.11$ & $69.57 \pm 1.49$ & $55.23 \pm 1.43$ & $53.37$ \\
2.7B FWE EN RPJ2 FR & $32.61 \pm 1.38$ & $49.76 \pm 1.05$ & $51.34 \pm 0.52$ & $67.25 \pm 1.09$ & $70.09 \pm 1.48$ & $54.73 \pm 1.43$ & $54.30$ \\
2.7B FWE EN RPJ2 FR 90\% & $33.65 \pm 1.40$ & $51.30 \pm 1.05$ & $52.61 \pm 0.52$ & $67.90 \pm 1.09$ & $71.77 \pm 1.46$ & $55.88 \pm 1.43$ & $55.52$ \\
\bottomrule
\end{tabular}
}
\caption{Evaluation of models on general understanding tasks for English (top) and French (bottom) with varying quality. Models are trained on FineWebEDU (FWE) in English and Redpajama2 (RPJ2) in French with and without filtering at varying model sizes. All evaluations are zero-shot.}
\label{tab:indiv_model_scaling_bi}
\end{table*}

\subsection{Comparisons for Public Models}

In this section, we provide individual accuracies for all tasks and models in Table~\ref{tab:fr_comp_public} and \ref{tab:3B_fr_comp_public}.  Table~\ref{tab:public_core} shows results for Core tasks. Table~\ref{tab:public_frenchbench} shows results for the FrenchBench multiple choice tasks. Table~\ref{tab:public_regional_nli} shows results for both the regional knowledge tasks, and the NLI tasks.

\begin{table*}
\resizebox{\textwidth}{!}{
    \begin{tabular}{lcccccc>{\columncolor{LightCyan}}c}
\toprule
Model Name & ARC-C & ARC-E & HS & PIQA & SCIQ & WG & AVG \\
\midrule
1.3B FWE EN RPJ2 FR & $35.32 \pm 1.40$ & $61.57 \pm 1.00$ & $52.56 \pm 0.50$ & $70.73 \pm 1.06$ & $81.60 \pm 1.23$ & $56.67 \pm 1.39$ & $59.74$ \\
1.3B FWE EN FW2 FR & $34.64 \pm 1.39$ & $62.25 \pm 0.99$ & $54.31 \pm 0.50$ & $72.31 \pm 1.04$ & $80.30 \pm 1.26$ & $57.77 \pm 1.39$ & $60.26$ \\
1.3B FWE EN FWE FR & $37.29 \pm 1.41$ & $64.06 \pm 0.98$ & $55.22 \pm 0.50$ & $73.45 \pm 1.03$ & $83.40 \pm 1.18$ & $57.70 \pm 1.39$ & $61.85$ \\
1.3B FWE EN RPJ 2 90\% & $34.90 \pm 1.39$ & $62.46 \pm 0.99$ & $52.75 \pm 0.50$ & $71.33 \pm 1.06$ & $80.70 \pm 1.25$ & $56.27 \pm 1.39$ & $59.73$ \\
1.3B FWE EN FW2 FR 90\% & $36.26 \pm 1.40$ & $62.37 \pm 0.99$ & $55.74 \pm 0.50$ & $73.50 \pm 1.03$ & $81.30 \pm 1.23$ & $57.22 \pm 1.39$ & $61.07$ \\
2.7B FWE EN RPJ2 FR & $38.65 \pm 1.42$ & $68.60 \pm 0.95$ & $57.82 \pm 0.49$ & $73.12 \pm 1.03$ & $84.20 \pm 1.15$ & $58.33 \pm 1.39$ & $63.45$ \\
2.7B FWE EN RPJ2 FR 90\% & $38.23 \pm 1.42$ & $66.08 \pm 0.97$ & $57.04 \pm 0.49$ & $73.18 \pm 1.03$ & $83.60 \pm 1.17$ & $59.19 \pm 1.38$ & $62.89$ \\
CroissantLLM & $27.56 \pm 1.31$ & $52.27 \pm 1.02$ & $53.53 \pm 0.50$ & $71.60 \pm 1.05$ & $79.40 \pm 1.28$ & $55.64 \pm 1.40$ & $56.67$ \\
TransWebLLM & $36.18 \pm 1.40$ & $62.21 \pm 0.99$ & $52.32 \pm 0.50$ & $70.51 \pm 1.06$ & $80.20 \pm 1.26$ & $56.27 \pm 1.39$ & $59.61$ \\
Bloom 1B & $25.68 \pm 1.28$ & $45.45 \pm 1.02$ & $42.98 \pm 0.49$ & $67.14 \pm 1.10$ & $74.20 \pm 1.38$ & $54.93 \pm 1.40$ & $51.73$ \\
Qwen2.5 1.5B & $45.14 \pm 1.45$ & $71.46 \pm 0.93$ & $67.75 \pm 0.47$ & $76.06 \pm 1.00$ & $92.90 \pm 0.81$ & $63.38 \pm 1.35$ & $69.45$ \\
EuroLLM 1.7B & $37.46 \pm 1.41$ & $64.10 \pm 0.98$ & $59.38 \pm 0.49$ & $73.45 \pm 1.03$ & $84.90 \pm 1.13$ & $59.04 \pm 1.38$ & $63.05$ \\
Helium-1 2B & $46.50 \pm 1.46$ & $73.74 \pm 0.90$ & $69.63 \pm 0.46$ & $78.62 \pm 0.96$ & $92.30 \pm 0.84$ & $66.77 \pm 1.32$ & $71.26$ \\
Bloom 3B & $30.55 \pm 1.35$ & $53.24 \pm 1.02$ & $54.51 \pm 0.50$ & $70.51 \pm 1.06$ & $81.70 \pm 1.22$ & $58.72 \pm 1.38$ & $58.21$ \\
Qwen2.5 3B & $47.44 \pm 1.46$ & $73.11 \pm 0.91$ & $73.53 \pm 0.44$ & $78.84 \pm 0.95$ & $93.80 \pm 0.76$ & $68.51 \pm 1.31$ & $72.54$ \\
\midrule 
1.3B FWE EN RPJ2 FR & $30.34 \pm 1.36$ & $47.16 \pm 1.05$ & $47.37 \pm 0.52$ & $64.58 \pm 1.12$ & $69.67 \pm 1.49$ & $53.83 \pm 1.43$ & $52.16$ \\
1.3B FWE EN FW2 FR & $31.21 \pm 1.37$ & $48.75 \pm 1.05$ & $50.27 \pm 0.52$ & $68.12 \pm 1.09$ & $70.30 \pm 1.48$ & $53.25 \pm 1.43$ & $53.65$ \\
1.3B FWE EN FWE FR & $33.39 \pm 1.39$ & $53.46 \pm 1.05$ & $47.68 \pm 0.52$ & $62.57 \pm 1.13$ & $72.40 \pm 1.45$ & $55.39 \pm 1.43$ & $54.15$ \\
1.3B FWE EN RPJ 2 90\% & $31.12 \pm 1.37$ & $49.45 \pm 1.05$ & $48.89 \pm 0.52$ & $65.94 \pm 1.11$ & $69.57 \pm 1.49$ & $55.23 \pm 1.43$ & $53.37$ \\
1.3B FWE EN FW2 FR 90\% & $32.61 \pm 1.38$ & $50.15 \pm 1.05$ & $53.21 \pm 0.52$ & $69.26 \pm 1.08$ & $71.14 \pm 1.47$ & $56.46 \pm 1.42$ & $55.47$ \\
2.7B FWE EN RPJ2 FR & $32.61 \pm 1.38$ & $49.76 \pm 1.05$ & $51.34 \pm 0.52$ & $67.25 \pm 1.09$ & $70.09 \pm 1.48$ & $54.73 \pm 1.43$ & $54.30$ \\
2.7B FWE EN RPJ2 FR 90\% & $33.65 \pm 1.40$ & $51.30 \pm 1.05$ & $52.61 \pm 0.52$ & $67.90 \pm 1.09$ & $71.77 \pm 1.46$ & $55.88 \pm 1.43$ & $55.52$ \\
CroissantLLM & $27.90 \pm 1.32$ & $45.22 \pm 1.04$ & $50.52 \pm 0.52$ & $66.87 \pm 1.10$ & $69.67 \pm 1.49$ & $55.31 \pm 1.43$ & $52.58$ \\
TransWebLLM & $34.79 \pm 1.41$ & $53.68 \pm 1.05$ & $48.21 \pm 0.52$ & $63.76 \pm 1.12$ & $75.45 \pm 1.39$ & $54.40 \pm 1.43$ & $55.05$ \\
Bloom 1B & $27.03 \pm 1.31$ & $40.03 \pm 1.03$ & $41.56 \pm 0.51$ & $61.70 \pm 1.13$ & $67.16 \pm 1.52$ & $54.73 \pm 1.43$ & $48.70$ \\
Qwen2.5 1.5B & $32.69 \pm 1.39$ & $51.12 \pm 1.05$ & $49.63 \pm 0.52$ & $63.06 \pm 1.13$ & $79.54 \pm 1.31$ & $57.86 \pm 1.42$ & $55.65$ \\
EuroLLM 1.7B & $31.39 \pm 1.37$ & $51.17 \pm 1.05$ & $51.47 \pm 0.52$ & $65.18 \pm 1.11$ & $74.08 \pm 1.42$ & $56.38 \pm 1.42$ & $54.94$ \\
Helium-1 2B & $36.09 \pm 1.42$ & $55.00 \pm 1.04$ & $59.51 \pm 0.51$ & $67.90 \pm 1.09$ & $81.64 \pm 1.25$ & $60.66 \pm 1.40$ & $60.13$ \\
Bloom 3B & $30.08 \pm 1.35$ & $45.49 \pm 1.05$ & $51.04 \pm 0.52$ & $65.13 \pm 1.11$ & $69.78 \pm 1.49$ & $54.24 \pm 1.43$ & $52.62$ \\
Qwen2.5 3B & $38.19 \pm 1.44$ & $55.70 \pm 1.04$ & $58.58 \pm 0.51$ & $65.18 \pm 1.11$ & $84.58 \pm 1.17$ & $63.46 \pm 1.38$ & $60.95$ \\
\bottomrule
\end{tabular}
}
\caption{Evaluation of our models against public models on Core ``general understanding'' tasks for English (top) and French (bottom).  All evaluations are zero-shot.}
\label{tab:public_core}
\end{table*}

\begin{table*}[ht]
\centering
\resizebox{\textwidth}{!}{
    \begin{tabular}{lcccc}
    \toprule
        Model Name & ARC-C & Grammar & HS & Vocab\\
        \midrule 
        1.3B FWE EN RPJ2 FR & $30.54 \pm 1.35$ & $82.35 \pm 3.51$ & $47.4 \pm 0.52$ & $80.67 \pm 3.64$ \\
        1.3B FWE EN FW2 FR & $31.05 \pm 1.35$ & $80.67 \pm 3.64$ & $50.26 \pm 0.52$ & $78.99 \pm 3.75$\\
        1.3B FWE EN FWE FR & $36.7 \pm 1.41$ & $68.07 \pm 4.29$ & $47.59 \pm 0.52$ & $65.55 \pm 4.37$\\
        1.3B FWE EN RPJ 2 90\% & $32.34 \pm 1.37$ & $80.67 \pm 3.64$ & $48.92 \pm 0.52$ & $77.31 \pm 3.86$\\
        1.3B FWE EN FW2 FR 90\% & $33.79 \pm 1.38$ & $82.35 \pm 3.51$ & $53.21 \pm 0.52$ & $77.31 \pm 3.86$ \\
        2.7B FWE EN RPJ2 FR & $31.82 \pm 1.36$ & $80.67 \pm 3.64$ & $51.31 \pm 0.52$ & $81.51 \pm 3.57$ \\
        2.7B FWE EN RPJ2 FR 90\% & $36.53 \pm 1.41$ & $84.87 \pm 3.3$ & $52.63 \pm 0.52$ & $78.99 \pm 3.75$\\
        CroissantLLM & $28.74 \pm 1.32$ & $78.15 \pm 3.8$ & $50.52 \pm 0.52$ & $78.15 \pm 3.8$\\
        TransWebLLM & $37.64 \pm 1.42$ & $67.23 \pm 4.32$ & $48.32 \pm 0.52$ & $58.82 \pm 4.53$\\
        Qwen2.5 1.5B & $36.44 \pm 1.41$ & $75.63 \pm 3.95$ & $49.69 \pm 0.52$ & $74.79 \pm 4.0$\\
        EuroLLM 1.7B & $33.7 \pm 1.38$ & $80.67 \pm 3.64$ & $51.34 \pm 0.52$ & $75.63 \pm 3.95$ \\
        Helium-1 2B & $39.52 \pm 1.43$ & $78.15 \pm 3.8$ & $59.61 \pm 0.51$ & $79.83 \pm 3.69$\\
        Bloom 3B & $31.14 \pm 1.35$ & $78.99 \pm 3.75$ & $51.0 \pm 0.52$ & $79.83 \pm 3.69$ \\
        Qwen2.5 3B & $40.03 \pm 1.43$ & $77.31 \pm 3.86$ & $58.59 \pm 0.51$ & $78.15 \pm 3.8$\\
     \bottomrule
    \end{tabular}
}
\caption{Evaluation of our models against public models on FrenchBench multiple choice  tasks for  French language.  All evaluations are zero-shot.}
\label{tab:public_frenchbench}
\end{table*}

\begin{table*}[ht]
\centering
\resizebox{\textwidth}{!}{
    \begin{tabular}{lcccc}
    \toprule
        Model Name & Include & Kaleidoscope & XNLI & French Topic NLI\\
        \midrule 
        1.3B FWE EN RPJ2 FR & $44.15 \pm 2.43$ & $19.82 \pm 1.44$ & $46.1 \pm 1.0$ & $33.33 \pm 1.93$\\
        1.3B FWE EN FW2 FR & $42.0 \pm 2.41$ & $21.0 \pm 1.48$ & $46.22 \pm 1.0$ & $33.33 \pm 1.93$\\
        1.3B FWE EN FWE FR & $37.71 \pm 2.37$ & $21.39 \pm 1.49$ & $48.11 \pm 1.0$ & $36.17 \pm 1.96$\\
        1.3B FWE EN RPJ 2 90\% & $45.35 \pm 2.43$ & $20.47 \pm 1.46$ & $48.84 \pm 1.0$ & $38.33 \pm 1.99$ \\
        1.3B FWE EN FW2 FR 90\% & $45.82 \pm 2.44$ & $19.82 \pm 1.44$ & $47.71 \pm 1.0$ & $33.33 \pm 1.93$\\
        2.7B FWE EN RPJ2 FR & $48.21 \pm 2.44$ & $20.87 \pm 1.47$ & $48.39 \pm 1.0$ & $33.33 \pm 1.93$\\
        2.7B FWE EN RPJ2 FR 90\% & $21.52 \pm 1.49$ & $48.07 \pm 1.0$ & $33.33 \pm 1.93$ \\
        CroissantLLM & $41.05 \pm 2.41$ & $19.29 \pm 1.43$ & $49.32 \pm 1.0$ & $33.33 \pm 1.93$\\
        TransWebLLM & $39.86 \pm 2.39$ & $19.42 \pm 1.43$ & $47.27 \pm 1.0$ & $33.5 \pm 1.93$ \\
        Bloom 1B & $36.04 \pm 2.35$ & $21.78 \pm 1.5$ & $46.71 \pm 1.0$ & $33.5 \pm 1.93$\\
        Qwen2.5 1.5B & $39.86 \pm 2.39$ & $23.23 \pm 1.53$ & $45.26 \pm 1.0$ & $41.83 \pm 2.02$ \\
        EuroLLM 1.7B & $44.87 \pm 2.43$ & $20.21 \pm 1.46$ & $47.51 \pm 1.0$ & $33.33 \pm 1.93$ \\
        Helium-1 2B & $47.49 \pm 2.44$ & $21.52 \pm 1.49$ & $50.8 \pm 1.0$ & $33.5 \pm 1.93$ \\
        Bloom 3B & $42.24 \pm 2.42$ & $20.87 \pm 1.47$ & $47.67 \pm 1.0$ & $33.33 \pm 1.93$ \\
        Qwen2.5 3B & $47.26 \pm 2.44$ & $25.98 \pm 1.59$ & $45.14 \pm 1.0$ & $59.5 \pm 2.01$\\
     \bottomrule
    \end{tabular}
}
\caption{Evaluation of our models against public models on Regional knowledge and NLI tasks for  French language.  All evaluations are zero-shot.}
\label{tab:public_regional_nli}
\end{table*}

\end{document}